\documentclass[10pt,twocolumn,letterpaper]{article}

\usepackage[pagenumbers]{cvpr} %

\usepackage{graphicx}
\usepackage{amsmath}
\usepackage{amssymb}
\usepackage{booktabs}
\usepackage{tabularray}
\usepackage{graphicx}
\usepackage{fancyhdr}
\usepackage{float}
\usepackage{ulem}
\usepackage{xcolor}
\usepackage[pagebackref=true,breaklinks=true,letterpaper=true,colorlinks,citecolor=citecolor,bookmarks=false]{hyperref}
\usepackage{textpos}
\usepackage[table]{colortbl}
\definecolor{citecolor}{RGB}{30,102,235}

\usepackage[capitalize]{cleveref}
\crefname{section}{Sec.}{Secs.}
\Crefname{section}{Section}{Sections}
\Crefname{table}{Table}{Tables}
\crefname{table}{Tab.}{Tabs.}
\def\etal{\emph{et al.}\@\xspace}
\def\cvprPaperID{***} %
\def\confName{CVPR}
\def\confYear{2023}
\begin{document}

\newcommand{\sx}[1]{\textcolor{cyan}{sx: #1}}
\newcommand{\bp}[1]{\textcolor{cyan}{bp: #1}}
\definecolor{deemph}{gray}{0.6}
\newcommand{\gb}{\rowcolor{gray!20}}

\newcommand\blfootnote[1]{\begingroup\renewcommand\thefootnote{}\footnote{#1}\addtocounter{footnote}{-1}\endgroup}

\title{Memory Encoding Model}

\author{%
  \textbf{Huzheng Yang} \quad \textbf{{James Gee}\footnote[2]{}} \quad \textbf{{Jianbo Shi}\footnote[2]{}}\\
  University of Pennsylvania\\
}

\twocolumn[{%
\renewcommand\twocolumn[1][]{#1}%
\maketitle
\vspace{-8mm}
\begin{center}
    \centering
    \captionsetup{type=figure}
    \includegraphics[width=0.98\linewidth]{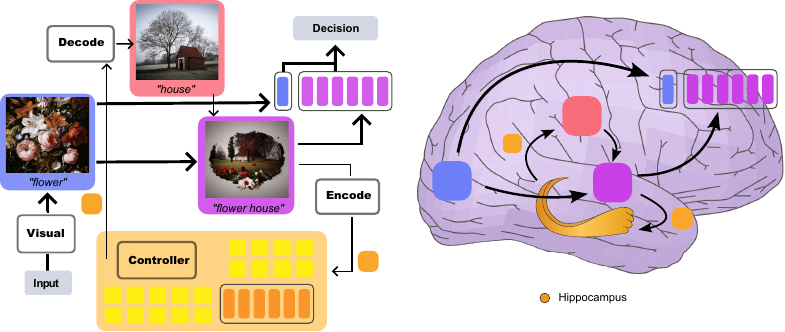}
    \vspace{0mm}
    \captionof{figure}{\textbf{Hypothetical memory replay process during the vision-memory task.} The task is to determine whether the current image was seen in the past. The visual brain is primarily involved in encoding the current image frame (blue) in to latent code. The non-visual brain is primarily decoding and regenerating memory frames (red). Latent code of current image (blue) is fused with memory frames (red), the fused image (purple) is stored in both working memory (purple) and hippocampus. Hippocampus maintain queues (orange) to control what to read from stored memory (yellow). The whole cortex is colored in purple to support the distributed nature of working memory.}\label{fig:theory}
\end{center}%
}]

\begin{abstract}
\vspace{-4mm}
  We explore a new class of brain encoding model by adding memory-related information as input. Memory is an essential brain mechanism that works alongside visual stimuli. During a vision-memory cognitive task, we found the non-visual brain is largely predictable using previously seen images. Our Memory Encoding Model (\texttt{Mem}) won the Algonauts 2023 visual brain competition even without model ensemble (single model score \textbf{66.8}, ensemble score \textbf{70.8}). Our ensemble model without memory input (\textbf{61.4}) can also stand a 3rd place.  Furthermore, we observe periodic delayed brain response correlated to 6th-7th prior image, and hippocampus also showed correlated activity timed with this periodicity. We conjuncture that the periodic replay could be related to memory mechanism to enhance the working memory . 
\end{abstract}

\blfootnote{Code and project page available \href{https://huzeyann.github.io/mem}{here}. \dag: co-advise.}
\section{Introduction}
\label{sec:intro}


\paragraph{Looking but not seeing, seeing but not looking.} The 
brain can be ``looking" at the image without ``seeing" it. The brain can think about an image without looking at it.  The brain can distinguish between novel and old images and encode them differently. None of the properties mentioned above are in the Vision Transformer (ViT) computer model.  Recent works made promising progress in modeling the visual brain based on one fundamental assumption: brain response is purely caused by the image presented at the current moment. Previous works with this assumption accurately predict the early visual cortex, but the complex and non-visual brain remains largely unpredictable.  

Our work in Algonauts 2023 visual brain competition leads to a curious, unexpected discovery of brain responses that suggest involuntary ``seeing'' without ``looking'' that persist throughout the brain region.  It is as if the brain constantly replays the images seen in a predictable periodicity without awareness.  
We reached this observation by examining the link between BOLD brain encoding to a variety of information beyond the current image. 


    \begin{enumerate}
        \vspace{-1mm}
        \item \textbf{Memory} state accumulated from previous images.
        \vspace{-1mm}
        \item \textbf{Behavior} response including button press and gaze.
        \vspace{-1mm}
        \item \textbf{Time} relative to the start of scanning session.
        \vspace{-1mm}
    \end{enumerate}

\begin{figure}
  \centering
  \begin{subfigure}[b]{0.45\textwidth}
    \includegraphics[width=\textwidth]{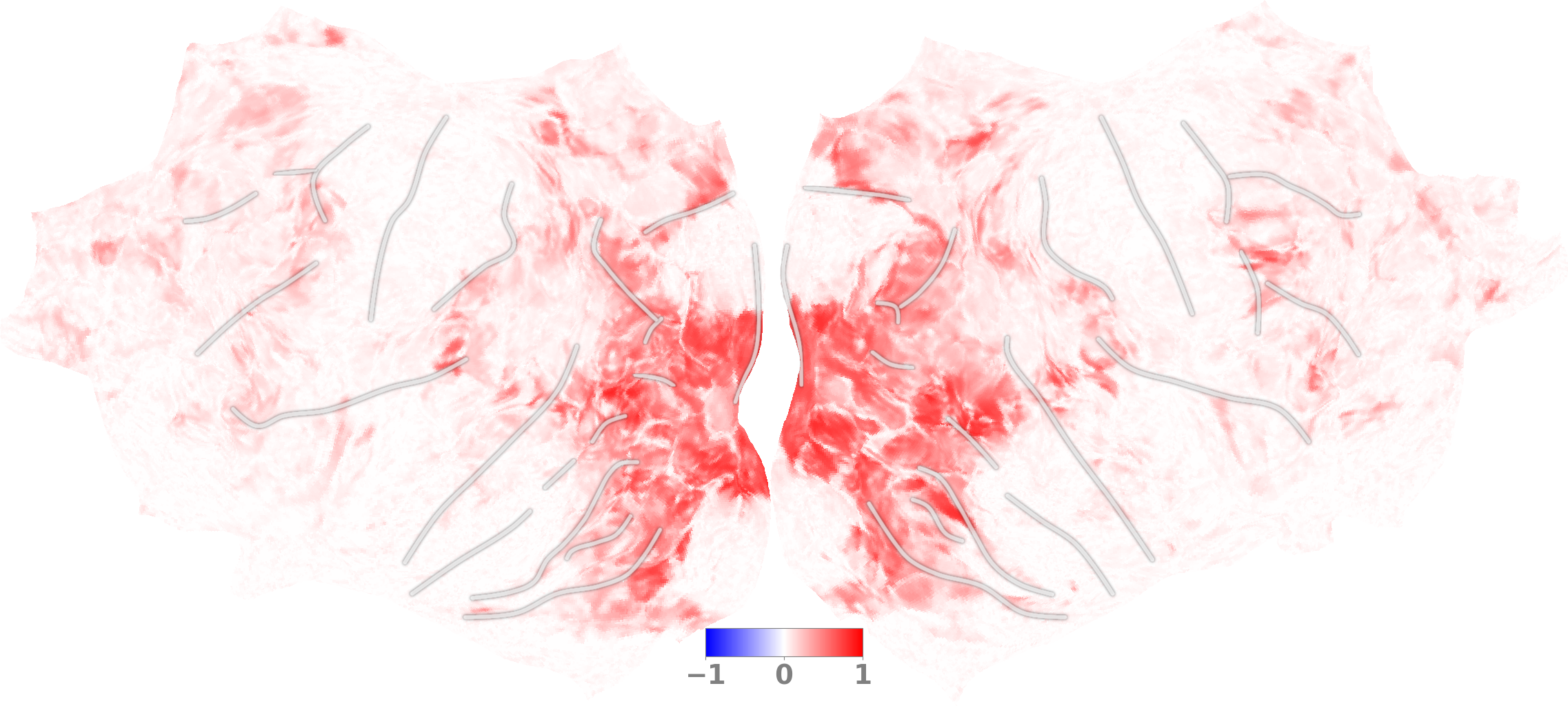}
    \caption{Noise Ceiling}
    \label{fig:baa}
  \end{subfigure}
  \hfill
  \begin{subfigure}[b]{0.45\textwidth}
    \includegraphics[width=\textwidth]{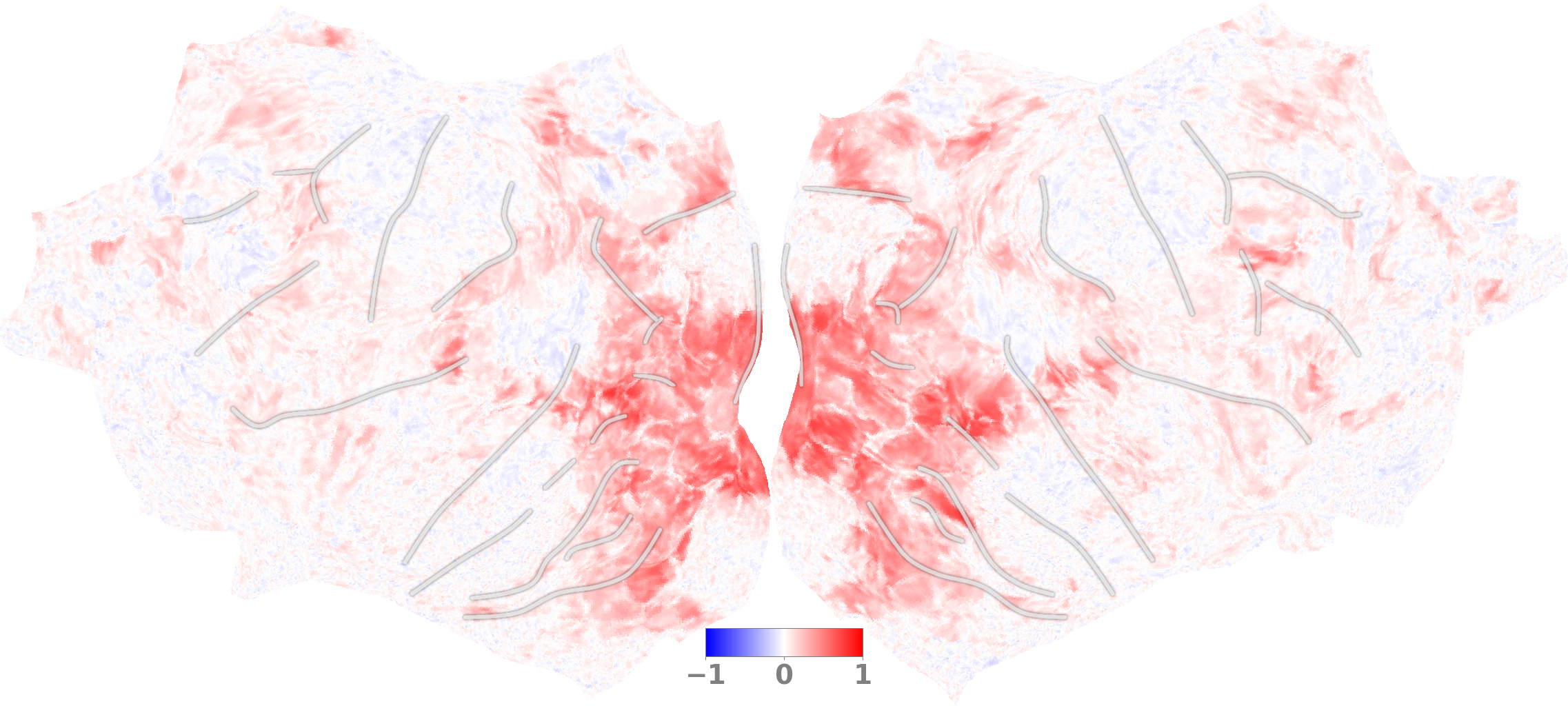}
    \caption{Baseline Model}
    \label{fig:bab}
  \end{subfigure}
  \hfill
  \begin{subfigure}[b]{0.45\textwidth}
    \includegraphics[width=\textwidth]{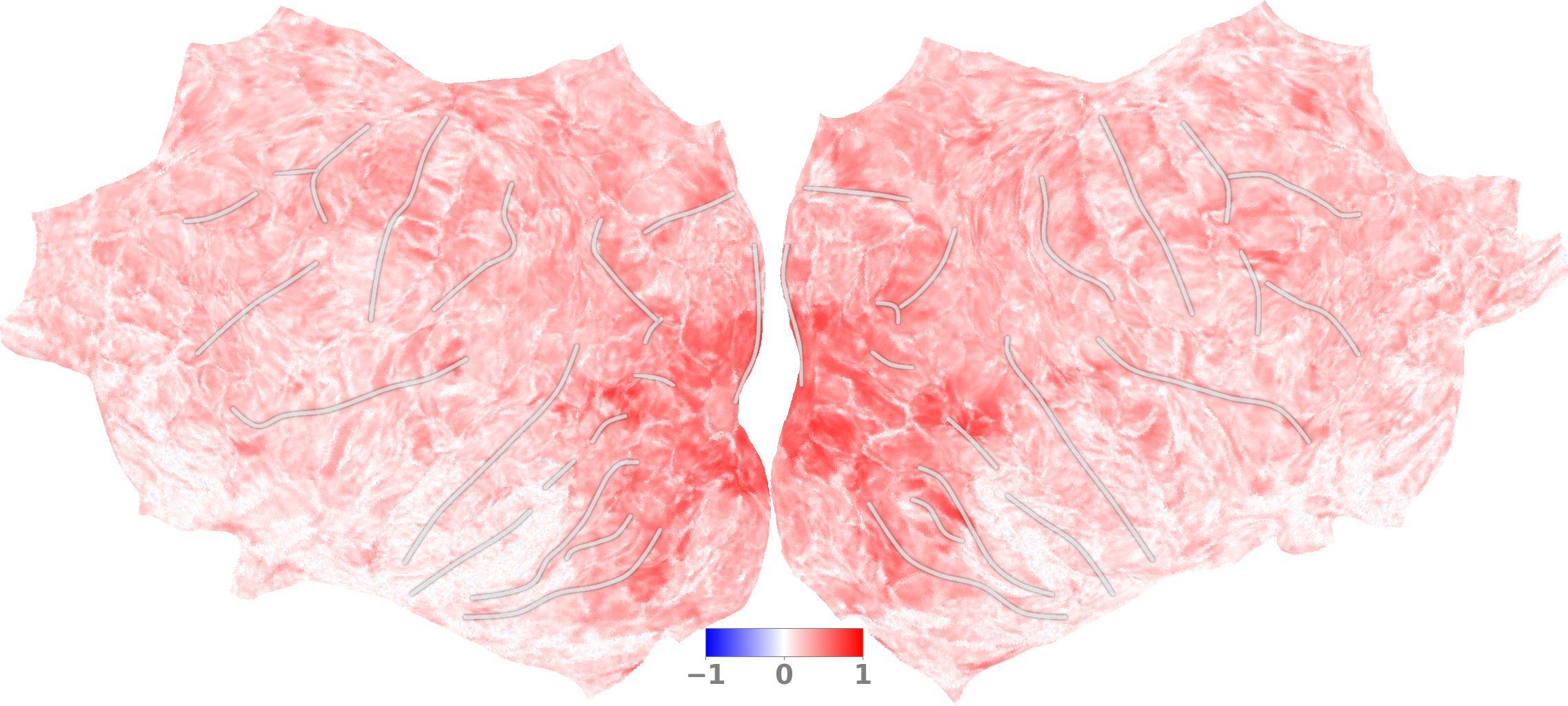}
    \caption{Full Mem Model}
    \label{fig:bac}
  \end{subfigure}
  \hfill
  \begin{subfigure}[b]{0.45\textwidth}
    \includegraphics[width=\textwidth]{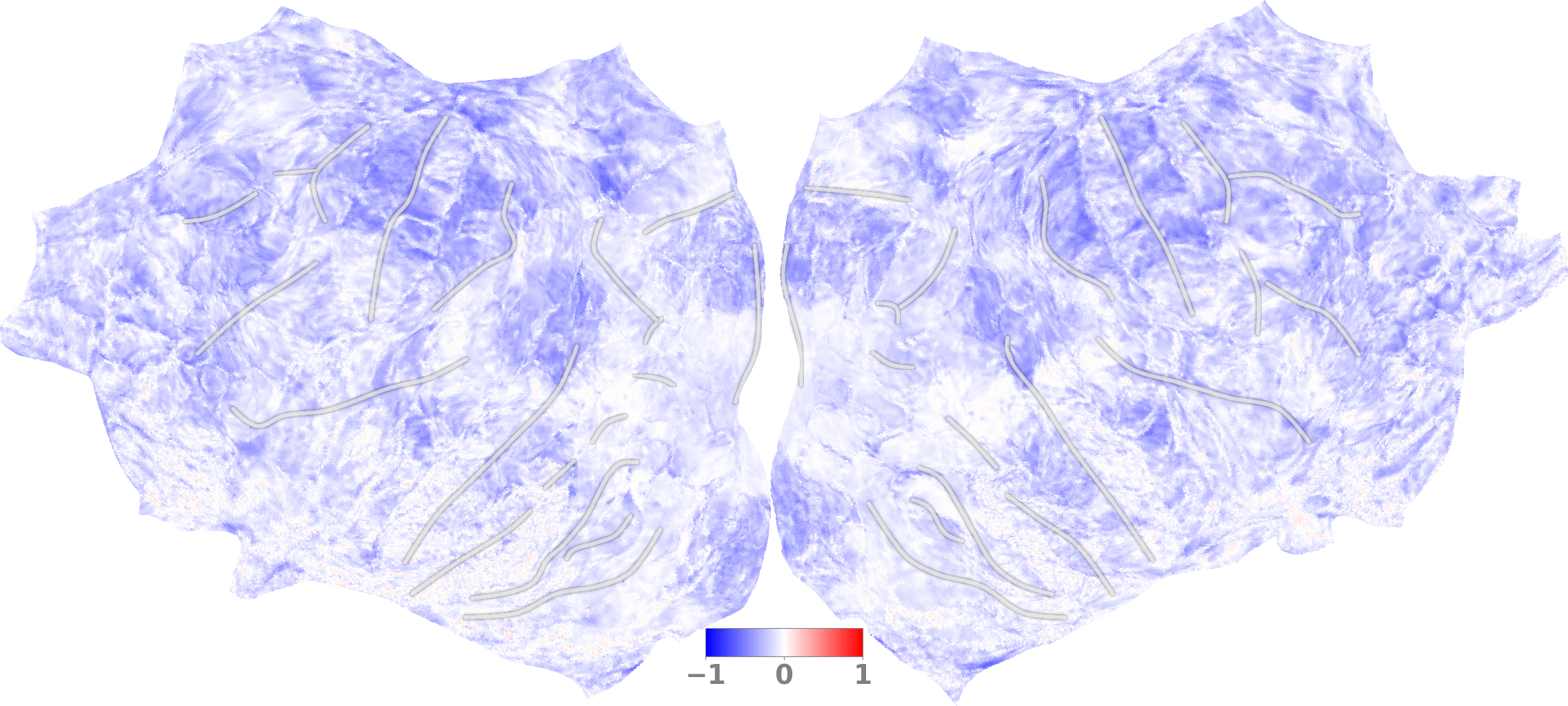}
    \caption{Difference: (b) - (c)}
    \label{fig:bad}
  \end{subfigure}
  
  \caption{\textbf{The whole cerebral cortex become largely predictable by considering memory.} Cortex plot shows prediction score (single-trial Pearson's $\mathbf{r}$). Prediction target is subject\#1 beta2 preparation. (a) Noise Ceiling measured by repeated trial. (b) Baseline model input is current image frame. (c) Full Mem model input are previous 32 frames, plus condition vector.}
  \label{fig:ba}
  \vspace{-2mm}
\end{figure}

\vspace{-3mm}
\paragraph{Whole brain is predictable.} We won the Algonauts 2023 visual brain competition with a score of 70.8, ahead of the 2nd place of 63.5 and the 5th place of 60.7.   We discover that when using memory-related information, previously seen images, the entire cortex becomes largely predictable (Figure \ref{fig:ba}), including the non-visual part.   In this challenge, the noise ceiling is nearly zero in the non-visual brain (Figure \ref{fig:baa}). However, our model score (Pearson's $\mathbf{r}$) is 0.16 in Somatomotor, 0.14 in Auditory, and 0.18 in Anterior cortex (Table \ref{tab:app_b23}). Noise ceiling treats the current image as ``signal" but other sources of information, including memory, as ``noise".  However, previously seen images contribute significantly to the increased prediction performance ($\Delta$score -4.15, Table \ref{tab:input}), and the contribution of behavioral responses are relatively minor ($\Delta$score -1.88).


\noindent Overall, two critical innovations led to our high-performance model: 

\begin{enumerate}
    \vspace{-1mm}
    \item \textbf{Modeling}: The model input combines the present image, behavioral response, and memory images. The \texttt{RetinaMapper} and \texttt{LayerSelector} module specifically focus on replicating retinotopy.
    \vspace{-2mm}
    \item \textbf{Training}: The random-ROI\footnote{\textit{ROI} refers to brain regions.} ensemble recipe benefits from implicit interactions and hierarchical organization of brain voxels. \vspace{-3mm}
\end{enumerate}

\vspace{-3mm}
\paragraph{Periodic delayed response.} 
We developed an ``information tracker" by building a prediction model that inputs previously seen images. With this tracker, we observe that responses in both the visual and non-visual brain (Figure \ref{fig:replay_cortex}) are correlated with the 6th-7th prior image but not the immediate past 1-2 image.  The correlation is periodic: we see a repeated response from the 12th-14th, up to the 32nd prior image.    The observation is consistent across all eight subjects in the NSD dataset.  The periodicity of 6-7 frames curiously matches that of working memory.  Furthermore, the hippocampus tail predicted response is to align with this periodicity, while the hippocampus head is not (Figure \ref{fig:replay_hippocampus}). This leads to our hypothesis (Figure \ref{fig:theory}): Memories are loaded from the hippocampus and regenerated periodically, perhaps to computationally enlarge the working memory. 

However, the same experiment conducted with the BOLD5000 dataset failed to find evidence of replay (Figure \ref{fig:bold5000}). Many uncontrolled factors exist between the two datasets, such as scanner field strength, image and blank presentation times, and the absence of a memory task. 
As both datasets use the same fMRI data preprocessing and denoising pipeline, this helps to invalidate the concern that replay is caused by systematic errors in data preprocessing: BOLD5000 didn't show what's suspicious to be `trend' (Figure \ref{fig:prevb2a} \ref{fig:bold5000}). Our theory of memory replay is yet to be explored through carefully controlled experiments.  We note that in auditory, we often experience involuntary replay, such as the ``earworm" phenomenon: speech or music loops itself in the brain. We cannot voluntarily forget memories or turn off earworms. Memory is an essential building block of the brain, and the study of visual replay should go alongside memory.


\begin{figure}
    \centering
    \includegraphics[width=0.47\textwidth]{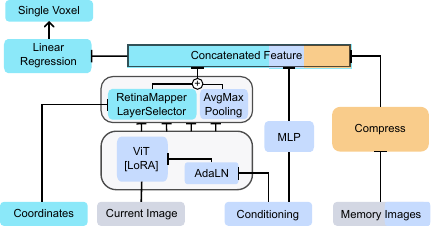}
    \caption{\textbf{Overview of the computational memory encoding model.} Model is illustrated for a \textbf{single voxel}, each voxel is uniquely predicted by: 1. unique physical coordinates that influence feature selection in \textit{RetinaMapper} and \textit{LayerSelector}. 2. voxel-specific linear regression weights.}
    \label{fig:model}
\end{figure}

\section{Related Work}
\label{sec:background}

Brain encoding model has been actively explored by the computational neuroscience community, highlighting from Kay \etal~\cite{kay_identifying_2008}, Naselaris \etal~\cite{naselaris_encoding_2011}, surveyed in \cite{wen_neural_2018}. The first step in the fMRI dataset is pre-processing and denoising, \cite{kay_glmdenoise_2013, prince_improving_2022} use a general linear model to denoise, significantly increase signal-to-noise ratio w.r.t. presented image.  Knowlege of brain Biological structure inspired the building of computational models. The early visual brain is doing retinotopy, from which the convolutional neural network was designed \cite{lecun_deep_2015}. Other works also introduce retinotopy mapping optimized for each voxel \cite{lurz_generalization_2021, st-yves_brain-optimized_2022}. Layers in feed-forward computer vision models are commonly reported to align with early visual to complex brain regions \cite{takagi_high-resolution_2022}. Besides single image models, recurrent neural network models is also studied \cite{khosla_cortical_2021} in video tasks.

\paragraph{Metrics and Benchmarks}
Representation similarity analysis (RSA) \cite{kriegeskorte_representational_2008} has been the most popular metric. Recent works prefer voxel-wise Pearson's $\mathbf{r}$ for a better resolution \cite{cichy_algonauts_2021,allen_massive_2022,willeke_sensorium_2022}. \cite{lurz_bayesian_2022} proposed a metric that measures full response distributions. There is a rising number of competition and benchmarks: the Brain-Score \cite{schrimpf_brain-score_2018} start with primate electrophysiology and extend to various source of data, the Algonauts \cite{cichy_algonauts_2021, gifford_algonauts_2023} on human fMRI, the Sensorium \cite{willeke_sensorium_2022} on mice calcium imaging. Other large-scale datasets include: the BOLD5000 \cite{chang_bold5000_2019}, the HCP 7T MOVIE \cite{van_essen_human_2012}, and the Things initiative \cite{hebart_things-data_2023,gifford_large_2022}. 

\section{Memory Encoding Model}
\label{sec:methods}

This section starts with a straightforward brain encoding model, and the input is only the current image frame. We add a retinotopy-inspired module called \texttt{RetinaMapper} and \texttt{LayerSelector}. Then we inject the conditioning vector into the image backbone by the \texttt{AdaLN} module. Lastly, we add input from memory images using the \texttt{Compressor} Module. An overview of the complete model is in Figure \ref{fig:model}.

\subsection{Straight-forward Model}
The goal is to build a model that predicts voxel-wise brain response $Y$ paired with input image $X$. It's a multi-task regression where each voxel is a task. We use a pre-trained image backbone model DiNOv2-ViT-B \cite{oquab_dinov2_2023} to build a straight-forward model:

\begin{equation}
\begin{aligned}
    M &= \texttt{ConvBlock}(\texttt{ViT}(X)) \\
    h &= \texttt{AvgMaxPool}(M) \\
    y_i &= w_i h + b_i \\
\end{aligned}
\label{eq:staight}
\end{equation}

\noindent
$M \in \mathcal{R}^{\frac{H}{k} \times \frac{W}{k} \times D'}$ is the feature map of one intermediate layer, \texttt{ConvBlock} is random initialized, $h \in \mathcal{R}^{1 \times 1 \times d}$. $i$ is the index of a voxel, and each voxel has non-weight sharing linear regression $w_i$ and $b_i$. LoRA optimizes attention layer, and MLP in ViT \cite{hu_lora_2021}. We found LoRA consistently gives a better score than frozen ViT, and completely unfreeze gives a lower score than frozen.

\subsection{RetinaMapper and LayerSelector}
The motivation is that spatially close voxels have similar functional roles, such as retinotopy mapping and face-selective regions. \texttt{RetinaMapper} and \texttt{LayerSelector} is designed to enforce the input feature $h$ to be unique for each voxel while being constrained by voxel's spatial coordinate $p_i \in \mathcal{R}^{3}$. 

\begin{figure}
  \centering

  \begin{subfigure}[b]{0.47\textwidth}
    \includegraphics[width=\textwidth]{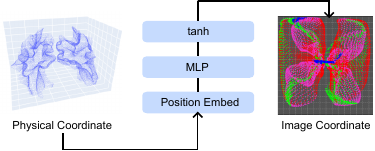}
    \caption{\texttt{RetinaMapper}}
    \label{fig:retinamappera}
  \end{subfigure}
  \hfill
  \begin{subfigure}[b]{0.47\textwidth}
    \includegraphics[width=\textwidth]{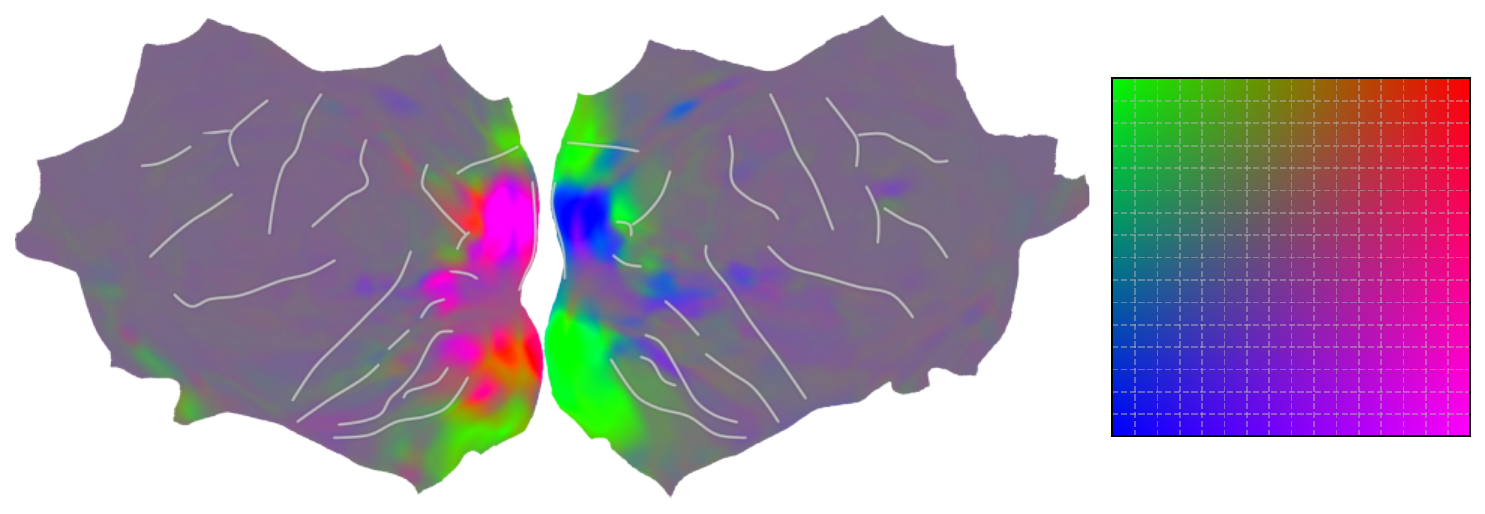}
    \caption{Retinotopic transform}
    \label{fig:retinamapperb}
  \end{subfigure}
\caption{\texttt{RetinaMapper} transform voxel's 3D physical coordinates into 2D latent image feature map space. The feature vector is sampled by linear interpolation at a 2D feature map; the same location is sampled for each hidden channel. (a) Image taken from \cite{yang_retinotopy_2023}, dots are colored by \texttt{argmax} of the \texttt{LayerSelector}, colors indicate selection of layers. (a) only includes visual brain voxels (nsdgeneral). (b) Retinotopic transformation is learned in end-to-end training without retinotopy supervision. The right side grid on (a) and (b) is the same latent image grid.}
\label{fig:retinamapper}
\end{figure}

\begin{figure}
  \centering

  \begin{subfigure}[b]{0.33\textwidth}
    \includegraphics[width=\textwidth]{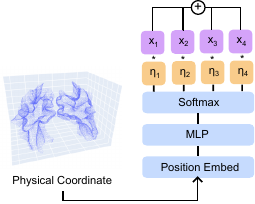}
    \caption{\texttt{LayerSelector}}
    \label{fig:layerselectora}
  \end{subfigure}
  \hfill
  \begin{subfigure}[b]{0.47\textwidth}
    \includegraphics[width=\textwidth]{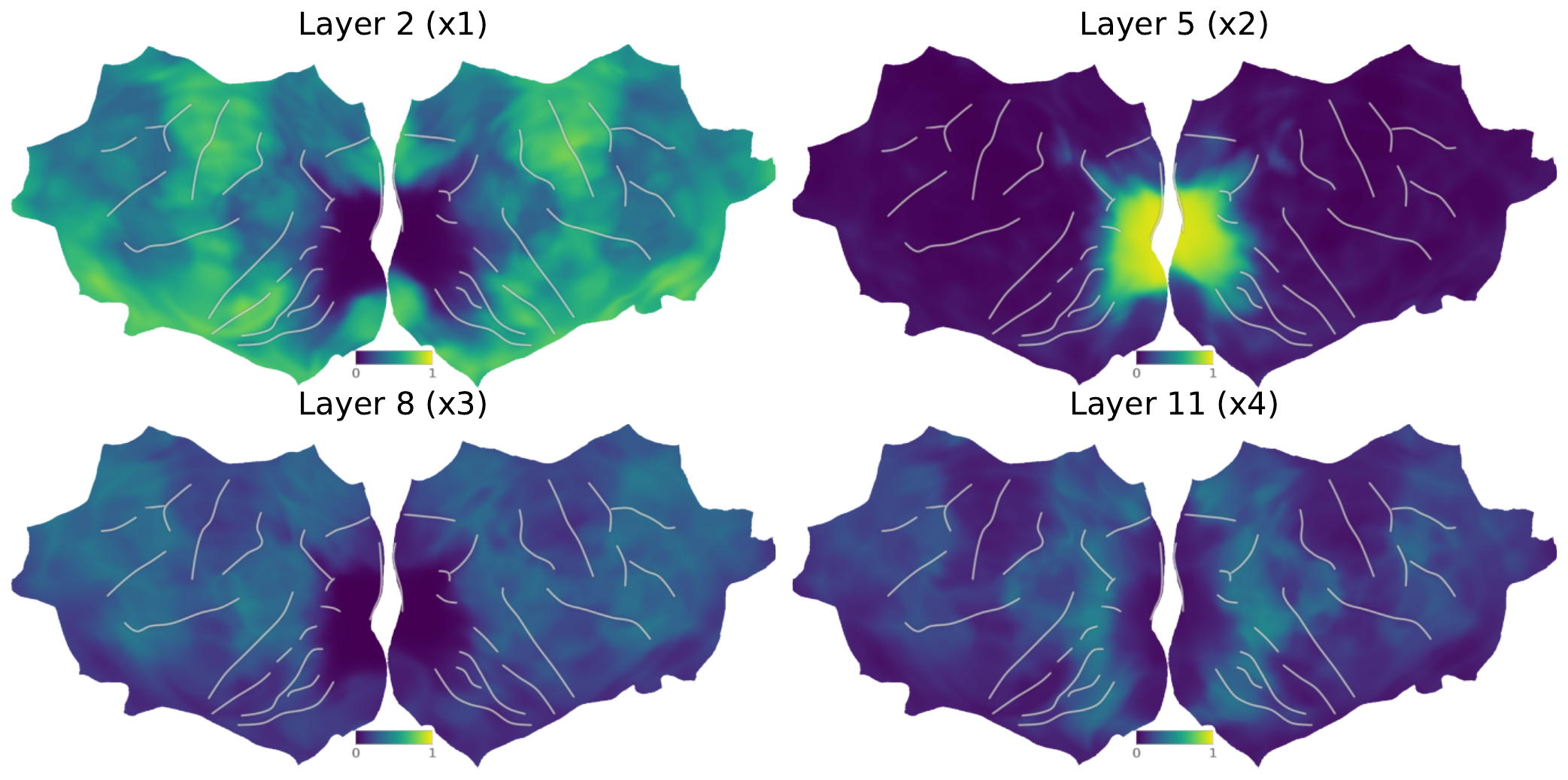}
    \caption{$\eta$}
    \label{fig:layerselectorb}
  \end{subfigure}
\caption{\texttt{LayerSelector} weights and sum feature vectors coming from various backbone layers. (a) $x_j$ is output from \texttt{RetinaMapper} thus unique for each voxel, $j$ is index for backbone layer. (b) $\eta$ showed for each backbone layers, vmax is 1.}
\label{fig:layerselector}
\end{figure}

\noindent
\paragraph{\texttt{RetinaMapper}} (Figure \ref{fig:retinamapper}) transforms voxel's spatial coordinate $p_i$ into image space 2D coordinate $u_i \in \mathcal{R}^{2}$
\begin{equation}
\begin{aligned}
    u = \texttt{tanh}(\texttt{MLP}(\texttt{PE}(p))) \\
\end{aligned}
\end{equation}
\noindent
\texttt{PE} is sinusoidal position encoding, \texttt{tanh} is necessary to make sure $u$ doesn't fall out-of-bound. The learnt \texttt{RetinaMapper} does follow the retinotopy motivation, with the exception of voxels stuck in the top left corner (Figure \ref{fig:retinamapperb} green). Note that \texttt{RetinaMapper} is only applied to current image and the stuck regions are not predictable by current image (Figure \ref{fig:bab}). 

\noindent
\paragraph{\texttt{LayerSelector}} (Figure \ref{fig:layerselector}) transforms voxel's spatial coordinate $p_i$ into ViT backbone layers weights $\eta_i \in \mathcal{R}^{4}$
\begin{equation}
\begin{aligned}
    \eta = \texttt{softmax}(\texttt{MLP}(\texttt{PE}(p))) \\
\end{aligned}
\end{equation}
\noindent
\texttt{softmax} makes $\eta_i$ sum to one. Entropy regularization ${L}_{ent}=\sum_{j=1}^{4} \eta^j \log \eta^j$ is applied to prevent converging to a local minimum in early training phase, $j$ is the index of 4 evenly spaced candidate layers.

\noindent
After adding \texttt{RetinaMapper} and \texttt{LayerSelector}, the simple model in equation \ref{eq:staight} become:
\vspace{-0mm}
\begin{equation}
\begin{aligned}
    M^j &= \texttt{ConvBlock}^j(\texttt{ViT}^j(X)) \\
    \check{h}^j &= \texttt{INTP}(M^j, \mathcal{\textbf{N}}(u, \sigma)) \\
    h^j &= \texttt{AvgMaxPool}(M^j) \\
    \tilde{h} &= \sum_{j=1}^{4} \eta^j (\check{h}^j + h^j) \\
    y_i &= w_i \tilde{h_i} + b_i \\
\end{aligned}
\end{equation}
\noindent
\texttt{INTP} is 2D linear interpolation on latent image grid $M^j \in \mathcal{R}^{\frac{H}{k} \times \frac{W}{k} \times D'}$, $\sigma = 0.01$ is fixed variance to reduce overfitting. $\tilde{h} \in \mathcal{R}^{\mathcal{N} \times d}$, $\mathcal{N}$ is number of voxels. Note that $\check{h}^j$ and $\tilde{h}$ is unique for each voxel. The learnt \texttt{LayerSelector} does follow the motivation that primary visual to complex regrions are matched from shallow to deep layer, with the exception of the first layer (Figure \ref{fig:layerselectorb} x1). Note that \texttt{LayerSelector} is only applied to current image and the first layer selected regions are not predictable by current image (Figure \ref{fig:bab}). 

\noindent
The same \texttt{RetinaMapper} is used for all layers. The effectiveness of \texttt{RetinaMapper} and \texttt{LayerSelector} is significant in visual brain and especially in primary visual (Table \ref{tab:model}).  

\subsection{Adaptive Layer Normalization}
\begin{figure}
    \centering
    \includegraphics[width=0.25\textwidth]{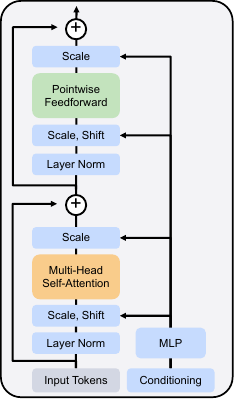}
    \caption{Adaptive Layer Normalization (AdaLN) module. Image taken from \cite{peebles_scalable_2023}}
    \label{fig:adaln}
\end{figure}

Figure \ref{fig:adaln}. The motivation is to inject the non-image conditioning vector $c \in \mathcal{R}^{d'}$ into the image backbone. Following \cite{peebles_scalable_2023}, we use the \texttt{AdaLN} module with shift and scale initialized to zero and one. Based on the observation that distribution heterogeneity exists on $c$ for different subject, we made the \texttt{AdaLN} module ``subject-aware":  
\begin{equation}
\begin{aligned}
    e &= \texttt{MLP}(\texttt{GeLU}(c w_s)) \\
    M &= \texttt{ConvBlock}(\texttt{AdaLNViT}(X; e)) \\
\end{aligned}
\end{equation}
\noindent
$w_s$ is a unique linear transform for each subject $s$. \texttt{MLP} is subject-shared. 

\newpage
\subsection{Memory Compressor}
\begin{figure}
    \centering
    \includegraphics[width=0.4\textwidth]{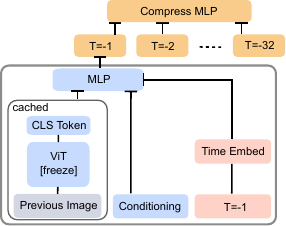}
    \caption{Previous image compression module. Images are cached to feature vectors, sinusoidal time embedding is concatenated as input, the same MLP (blue) is used for all previous images. Bottom part for $T<=-2$ is omitted.}
    \label{fig:compress}
\end{figure}

Figure \ref{fig:compress}. The input is both images $X$ and conditioning vectors $c$ from previous scan. Based on the experiment result that previous frame is osculating in a period, we designed the \texttt{Compressor} module to be "time-aware":
\begin{equation}
\begin{aligned}
    q_t &= \texttt{ViT}(X_t) \\
    \check{t} &= \texttt{TE}(t)w_s \\
    e_t &= \texttt{MLP}(\texttt{GeLU}(c_t {w'}_s)) \\
    \bar{q}_t &= \texttt{MLP}(q_t, e_t, \check{t}) \\
    \bar{h} &= \texttt{MLP}(\bar{q}) \\
\end{aligned}
\end{equation}
\noindent
$t$ is the index of previous frames, $q_t \in \mathcal{R}^{D'}$ is ViT class token, \texttt{TE} is sinusoidal time embedding. 

\subsection{Training Recipe}
Dividing voxels into ROIs loses the potential to aggregate knowledge and collaborate between ROIs. Mixing can also negatively affect individual voxel performance, making learning more challenging \cite{yang_retinotopy_2023}. 
Our \textit{random-ROI ensemble} recipe aims to gather the benefits of dividing and mixing. We train a zoo of models with different atlas ROI configurations (Table \ref{tab:recipe_method}). We then average each voxel's prediction from the model zoo.

\noindent
Inspired by \texttt{ModelSoup} \cite{wortsman_model_2022}, we greedily average the top 10 model checkpoint during training, this make each single model already an ``ensemble". However, at the time of writing, we do not have experiment evidence to support \textit{random-ROI ensemble} works better than a naive ensemble that use a fixed atlas.

\begin{table}
\centering
\begin{tblr}{
  cell{1}{2} = {c=4}{},
  cell{1}{6} = {c=2}{},
  hline{1,5} = {-}{0.08em},
  hline{3} = {-}{0.05em},
  colsep = 4pt, 
}
\textbf{Recipe} & \textbf{Number of} &              &                                 &                & \textbf{Parameters}             &               \\
                & \textbf{Atlas}     & \textbf{ROI} & {\textbf{Back-}\\\textbf{bone}} & \textbf{Voxel} & {\textbf{Back-}\\\textbf{bone}} & \textbf{Head} \\
Naive           & $1$                  & $1$            & $1$                               & $\mathcal{N}$              & $\mathcal{D}$                               & $\mathcal{N}d$            \\
Ensemble        & $a$                  & $b$            & $ab$                              & $a\mathcal{N}$             & $(ab)\mathcal{D}$                             & $a(\mathcal{N}d)$           
\end{tblr}
\captionsetup{font=normalsize} 
\caption{The ensemble recipe train models with various atlas configurations.}
\label{tab:recipe_method}
\end{table}

\section{Experiments}
\label{sec:experiments}

\subsection{Datasets}

\paragraph{NSD} \cite{allen_massive_2022} features large-scale and high signal-to-noise ratio 7T fMRI scan, the participants are asked to judge if the image was seen in the past. The image is presented for 3 sec followed by 1 sec blank; each run lasts for 300 sec. We use previous images from the same run (including blank) as memory images; each run is padded to the left with blank images. Conditioning vector is extracted as 1. \textbf{Memory}: statistics about when is the current image previously presented. 2. \textbf{Behavior}: button press and reaction time. 3. \textbf{Time}: relative time to the start of the session and the run. ROI for results reporting are HCPMMP1 \cite{glasser_multi-modal_2016} in fsaverage space. NSD provides hippocampus segmentation in volumetric space.

\paragraph{BOLD5000} \cite{chang_bold5000_2019} use same data pre-processing and denoising pipeline \cite{prince_improving_2022} as \textbf{NSD}. However, many complex factors contribute to a lower signal-to-noise ratio, the most obvious difference being a lower scanner field at 3T.  Most of BOLD5000 images are not repeatedly presented. The participants are asked to judge if they like the presented image. Image is presented for 1 sec followed by 9 sec blank, and each run lasts for 370 sec. 


\subsection{Metrics}

\paragraph{Noise ceiling is limited to early visual brain.}
The concept ``noise ceiling" \cite{allen_massive_2022} treats current image-unrelated signals as noise. In the GLMsigle denoising pipeline, beta3 preparation actively removes T=0 unrelated signal as `noise' by using a noise regressor; beta2 does have noise regressor. The noise ceiling is shown to be higher in beta3 than in beta2. However, we found that if previous images are added as input, the prediction score is higher in beta2 than beta3 (Table \ref{tab:app_b23}). Our model with memory input masked out shows a prediction score of a similar shape to the noise ceiling (Figure \ref{fig:baa} \ref{fig:bab}). Our model with memory input shows a score far beyond the noise ceiling in auditory, motor, and frontal areas. However, additional memory input showed little improvement in the early visual cortex (Figure \ref{fig:bad}). This suggests noise ceiling on the early visual brain is still an accurate measure of data quality.

\begin{figure}
\vspace{3mm}
  \centering
  
  \begin{subfigure}[b]{0.45\textwidth}
    \includegraphics[width=\textwidth]{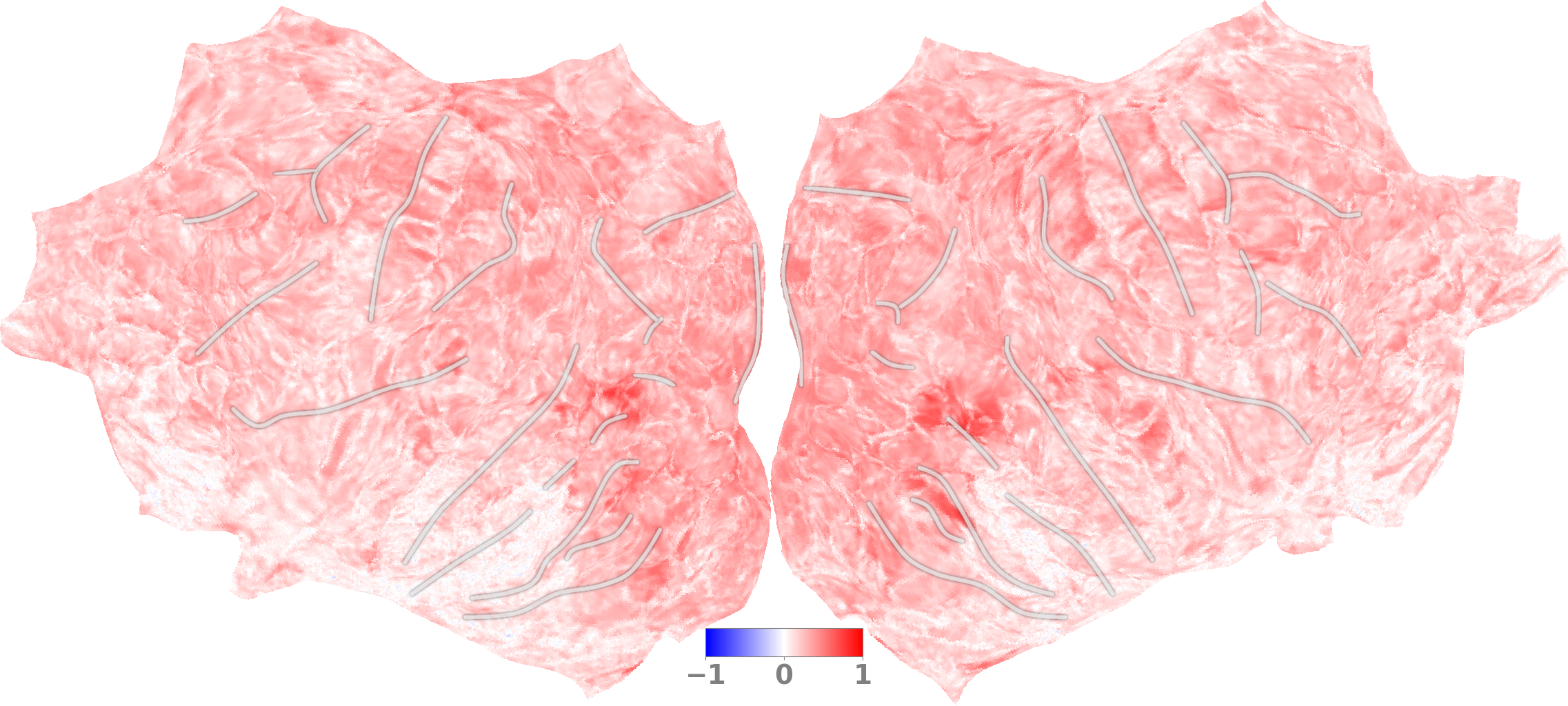}
    \caption{beta2}
    \label{fig:b3a}
  \end{subfigure}
  \hfill
  \begin{subfigure}[b]{0.45\textwidth}
    \includegraphics[width=\textwidth]{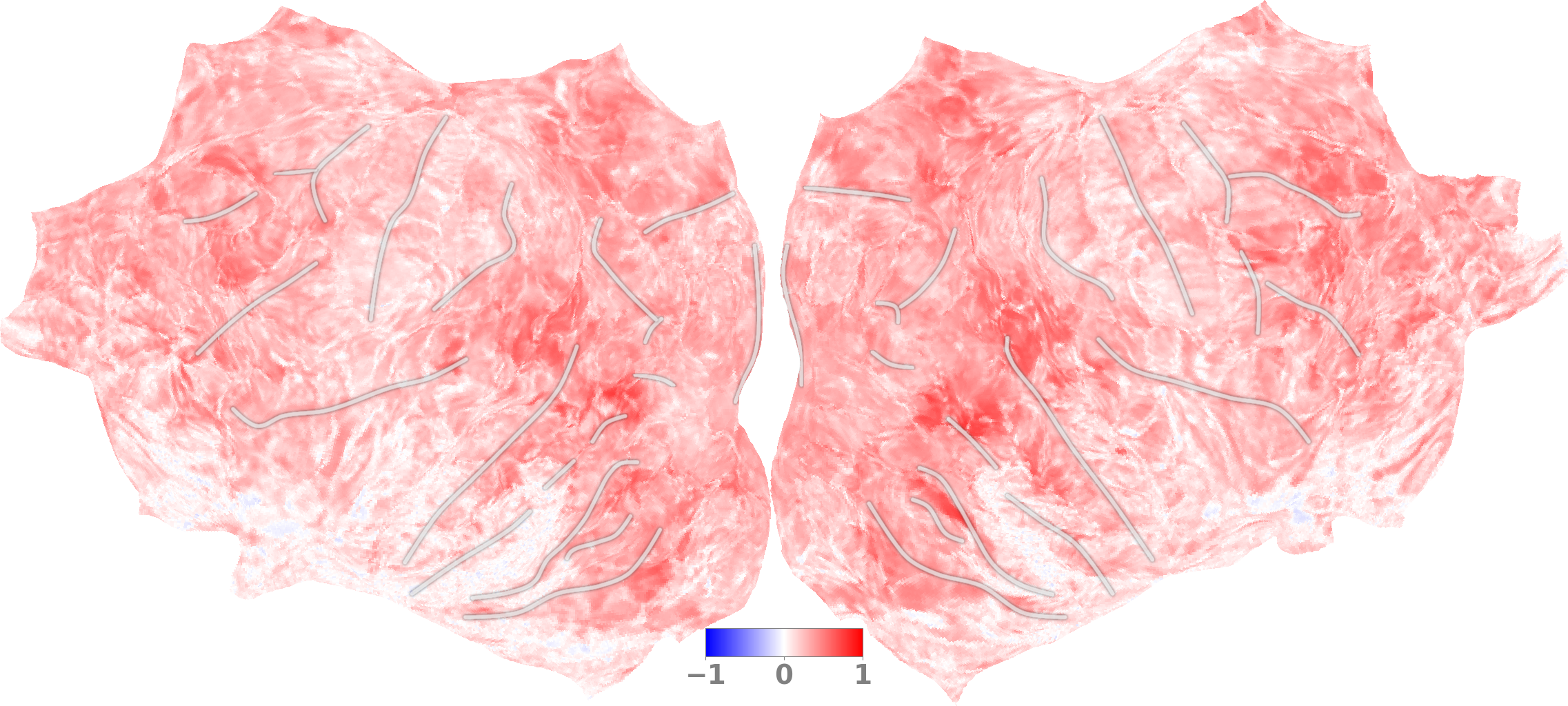}
    \caption{beta3}
    \label{fig:b3b}
  \end{subfigure}
  
  \caption{Prediction score, input is 32 image frames without conditioning vector. Results are single-trial Pearson's $\mathbf{r}$, showed for subject\#1, for other subjects please see online video. Note: performance is restricted by model size compared to Figure \ref{fig:ba}.}
  \label{fig:allframes}
\vspace{3mm}
\end{figure}

\begin{figure}
  \centering
  
  \begin{subfigure}[b]{0.45\textwidth}
    \includegraphics[width=\textwidth]{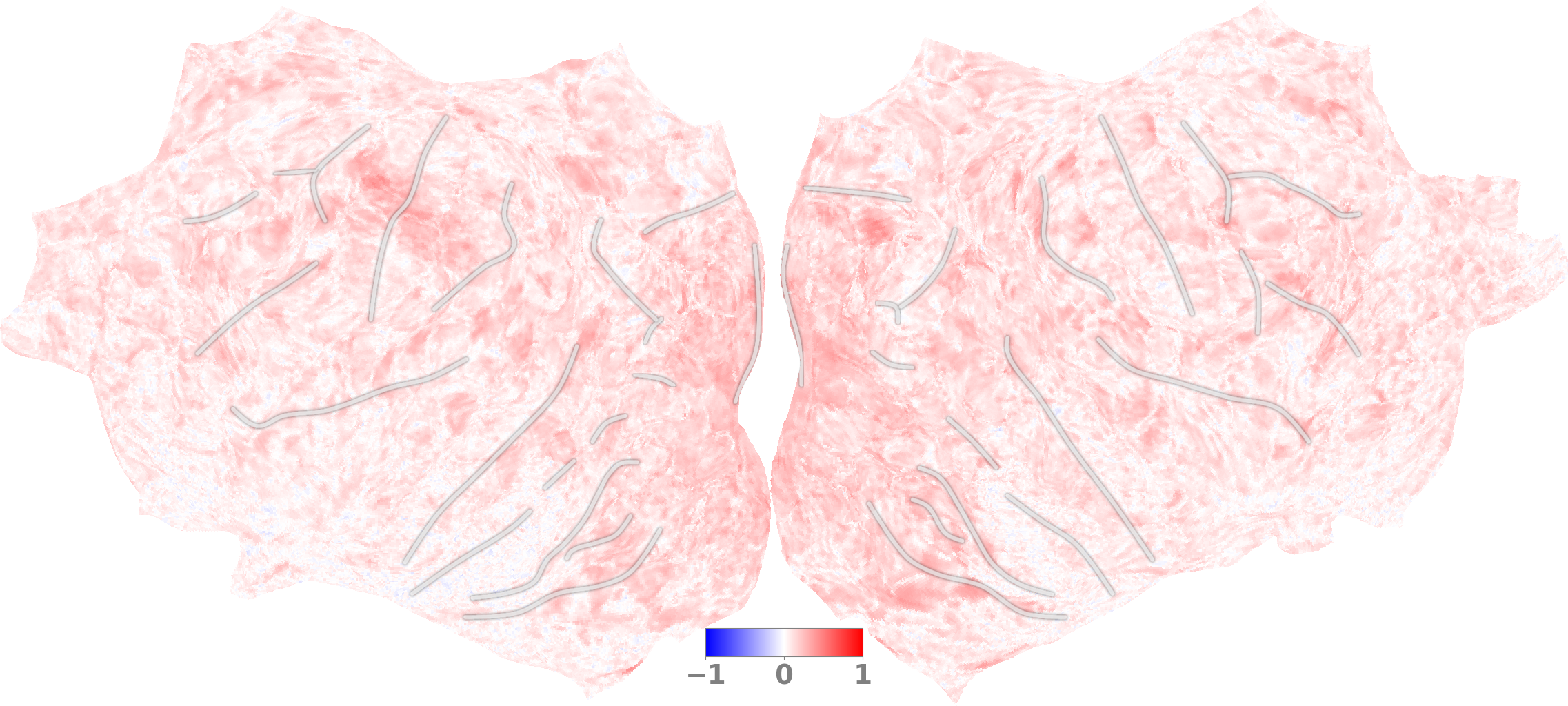}
    \caption{Input: time (run\_id and trial\_id)}
    \label{fig:c3a}
  \end{subfigure}
  \hfill
  \begin{subfigure}[b]{0.45\textwidth}
    \includegraphics[width=\textwidth]{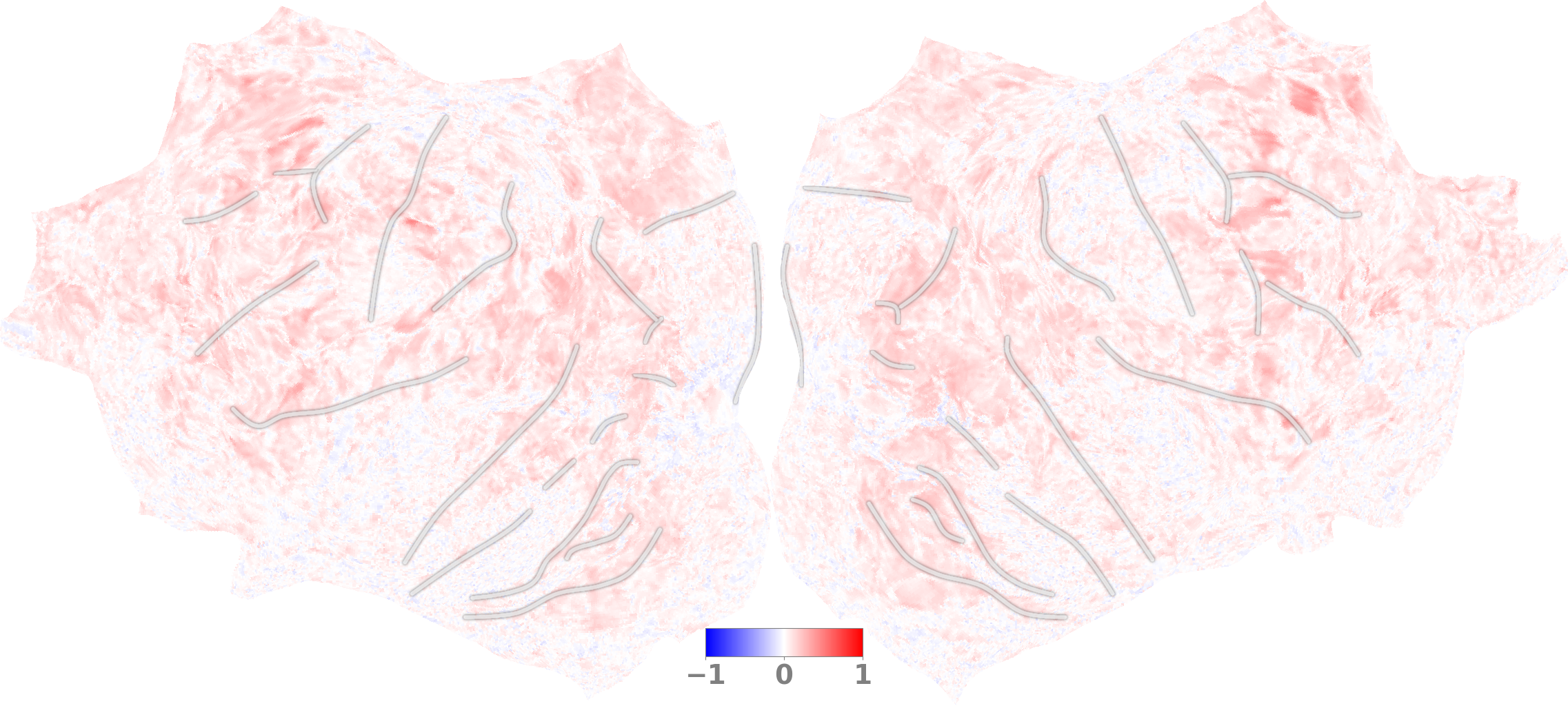}
    \caption{Input: button press}
    \label{fig:c3b}
  \end{subfigure}
  \hfill
  \begin{subfigure}[b]{0.45\textwidth}
    \includegraphics[width=\textwidth]{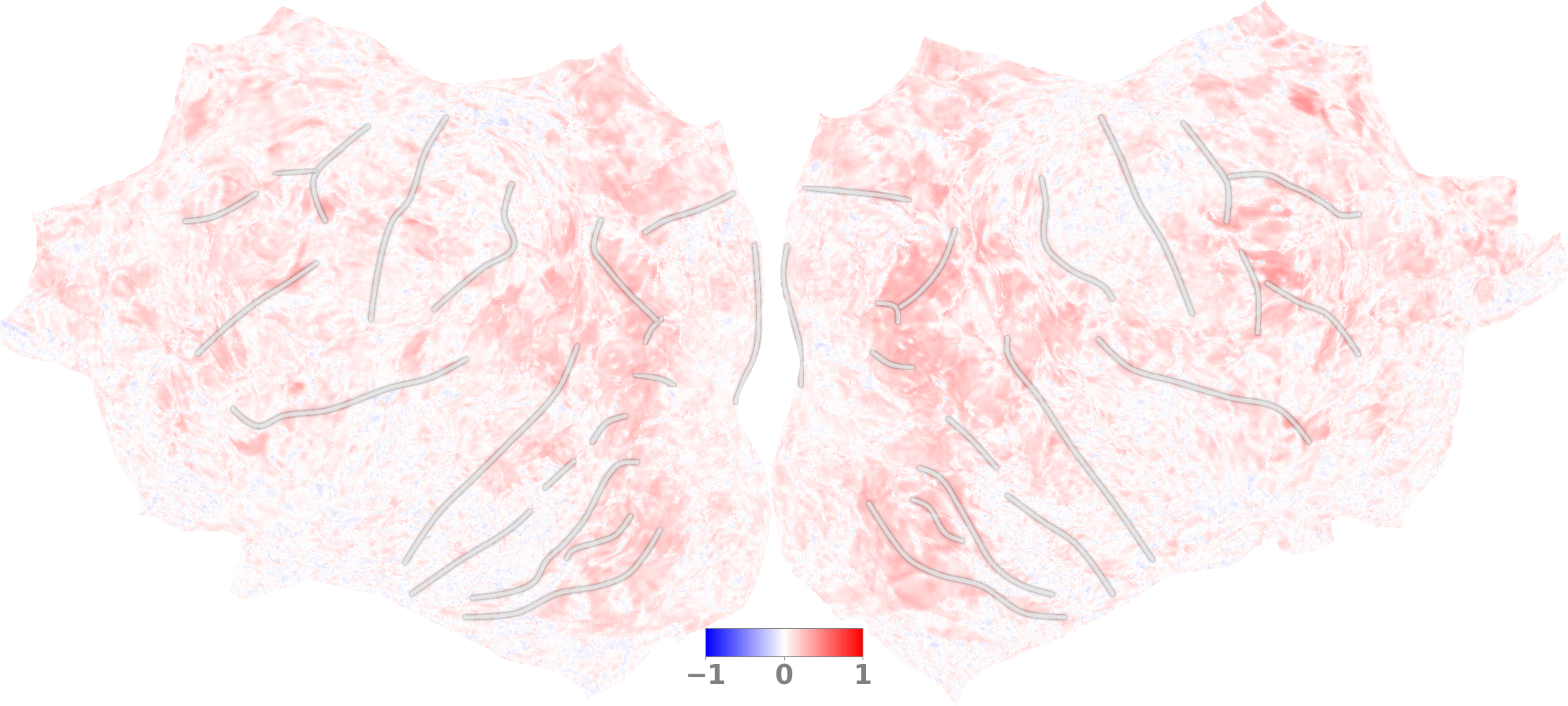}
    \caption{Input: old flags}
    \label{fig:c3c}
  \end{subfigure}
  \hfill
  \begin{subfigure}[b]{0.45\textwidth}
    \includegraphics[width=\textwidth]{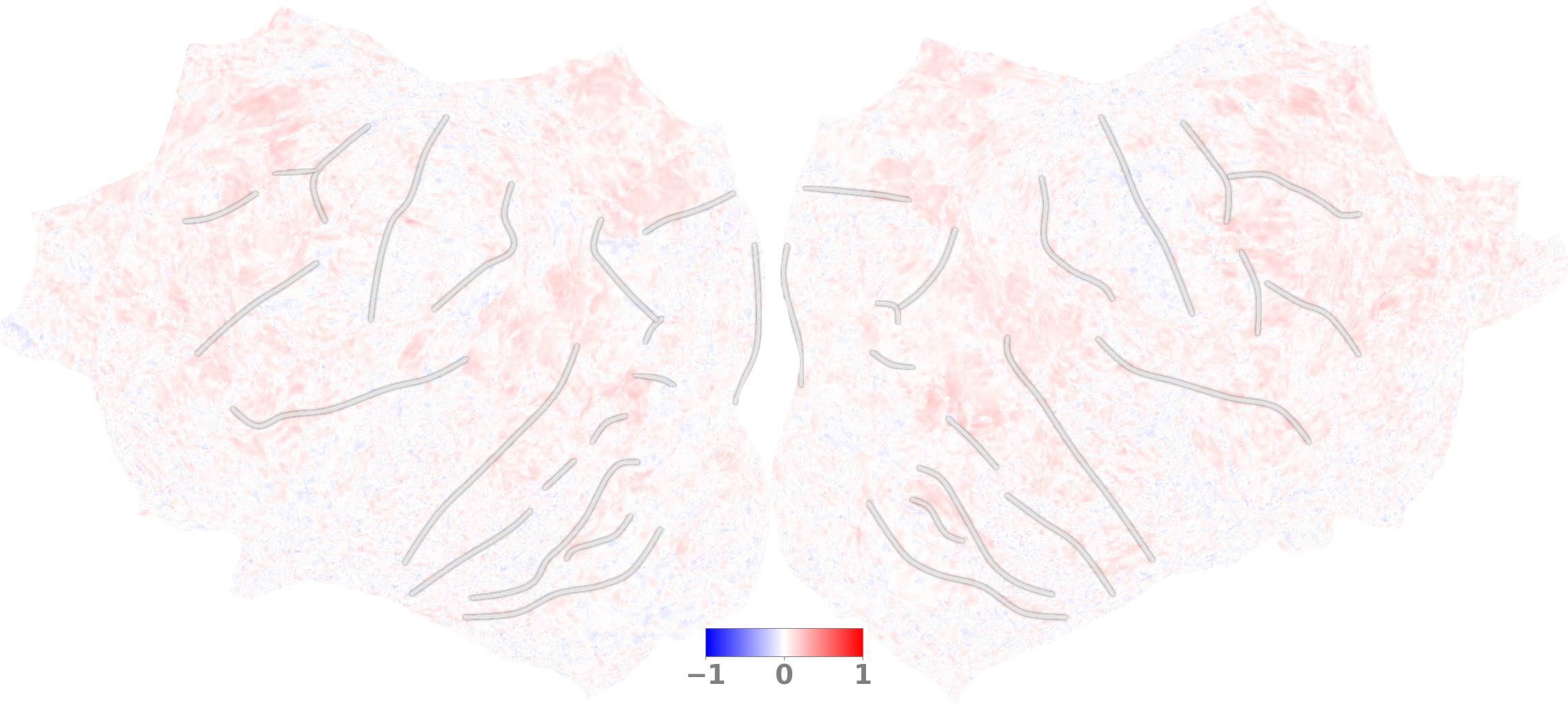}
    \caption{Input: future answer}
    \label{fig:c3d}
  \end{subfigure}
  \hfill
  \begin{subfigure}[b]{0.45\textwidth}
    \includegraphics[width=\textwidth]{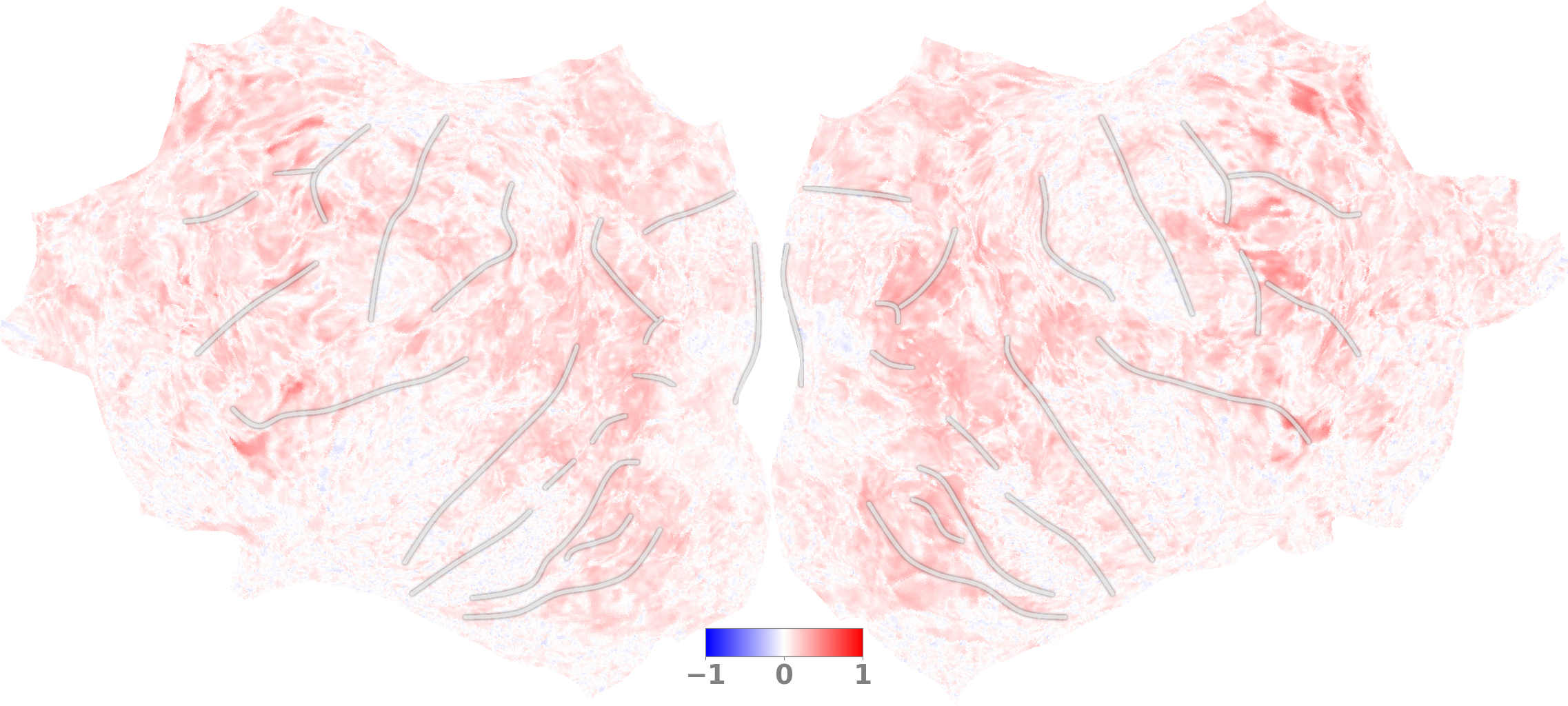}
    \caption{Input: reaction time}
    \label{fig:c3e}
  \end{subfigure}
  
  \caption{Prediction score without image input. Results are single-trial Pearson's $\mathbf{r}$, subject\#1 beta3, for other subjects please see online video.}
  \label{fig:cond_b3}
\end{figure}

\paragraph{Model score boosted by various input.}
Memory information boosts The model score from 58.82 to 66.80 (Table \ref{tab:input}). The increase in prediction power is majorly caused by image input from previous frame (Table \ref{tab:input}, Figure \ref{fig:allframes}). Memory information, when was the image seen last time, in conditioning vector only has a minor contribution that is within standard deviation (66.80 vs 67.02, Table \ref{tab:input}, Figure \ref{fig:c3c}). Behavior response input, including current answer, future answer, and reaction time, also shows a significant contribution (66.80 vs 64.92, Table \ref{tab:input}, Figure \ref{fig:c3b} \ref{fig:c3d} \ref{fig:c3e}). Time relative to the start of the session has a considerable prediction power when other information is removed (Figure \ref{fig:c3a}). However, its contribution is low (66.80 vs 66.09, Table \ref{tab:input}).

\paragraph{Model score boosted by ensemble and distillation.}
The random-ROI ensemble recipe boosted the model score from 66.80 to 70.85 (Table \ref{tab:input}) when with memory input, 58.82 to 61.39 when without memory input. We also explored model distillation. A distilled model can not perform as well as an ensemble (66.80 vs 70.85). Distillation from a naive recipe has a similar effect to distilling from an ensemble recipe (58.22 vs 58.82).

\subsection{Memory Replay}

\paragraph{Tracker} 
We build ``information tracker" by a prediction model.  Each model is a single-subject whole-brain. There are 32 frames, 8 subjects, 2 beta versions, and a Cartesian product 512 models are trained. For fast computation, the model size is restricted.

\begin{figure}
    \centering
    \includegraphics[width=0.45\textwidth]{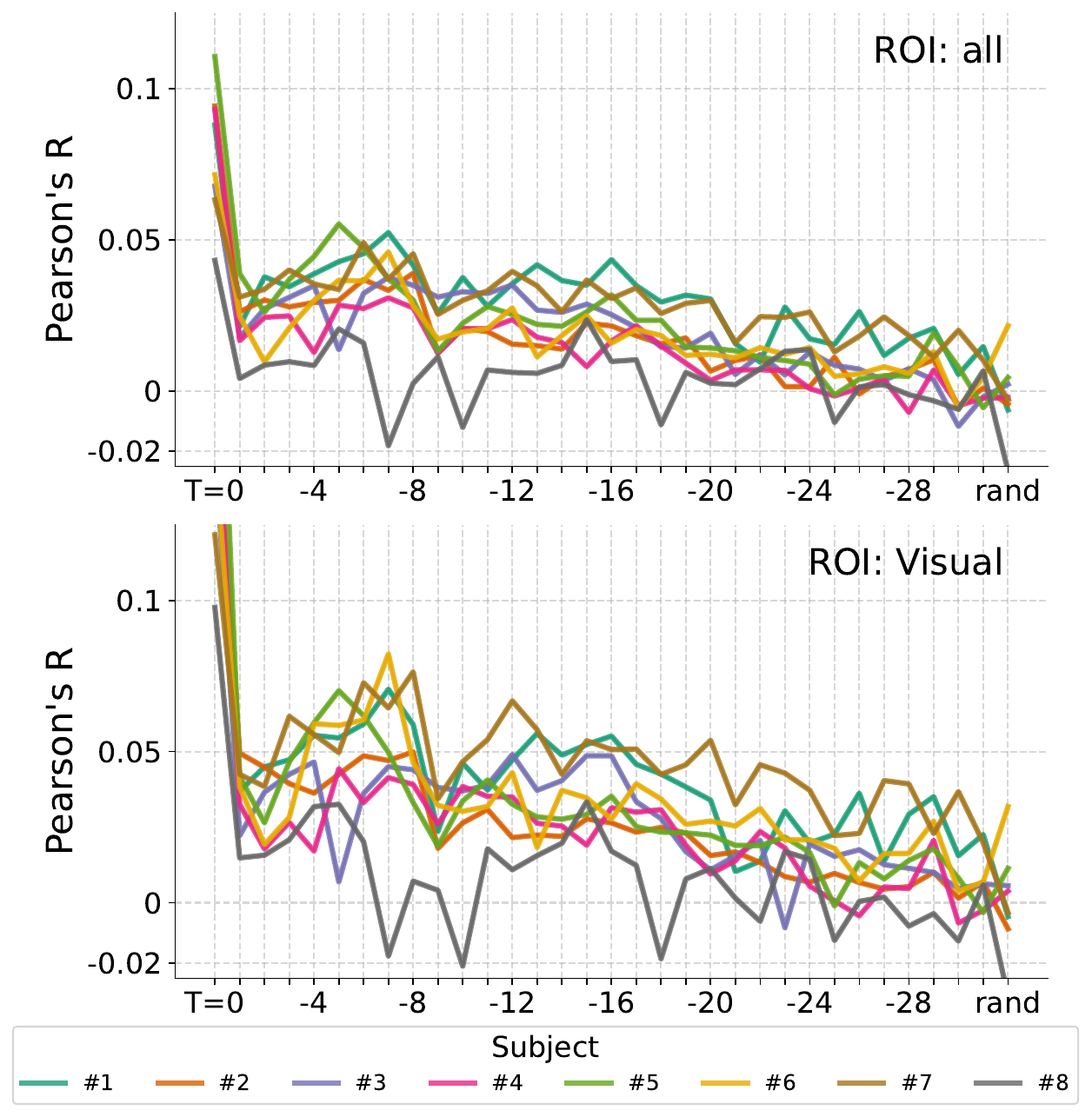}
    \caption{\textit{Periodic delayed response of previously seen images.} Periodic increase (period 6-8) can be observed at $T < -2$. The x-axis is index of previous image, y-axis is prediction score. Models are trained with each $T=-n$ as input. Results are beta3 preparation.}
    \label{fig:replay_cortex}
\end{figure}

\begin{figure}
    \centering
    \includegraphics[width=0.45\textwidth]{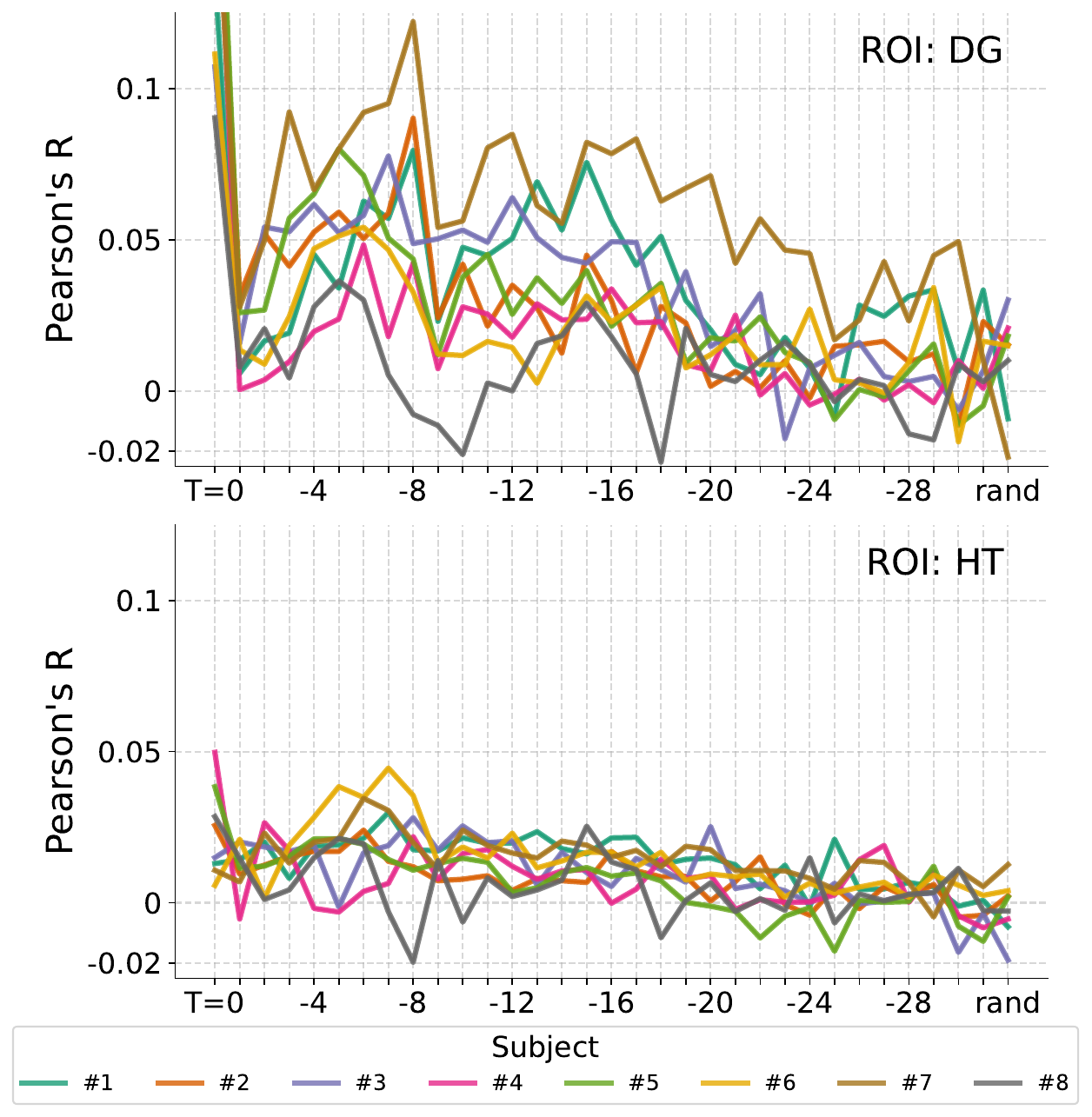}
    \caption{\textit{Read-write operation on hippocampus.} DG (Dentate Gyrus) is hippocampus head. HT is hippocampus tail. Hippocampus tail shows a peak higher than $T=0$ at $T=-6$, suggesting read operation on memory. The x-axis is index of previous image, y-axis is prediction score. Models are trained with each $T=-n$ as input. Results are beta3 preparation.}
    \label{fig:replay_hippocampus}
\end{figure}

\begin{figure}
    \centering
    \includegraphics[width=0.4\textwidth]{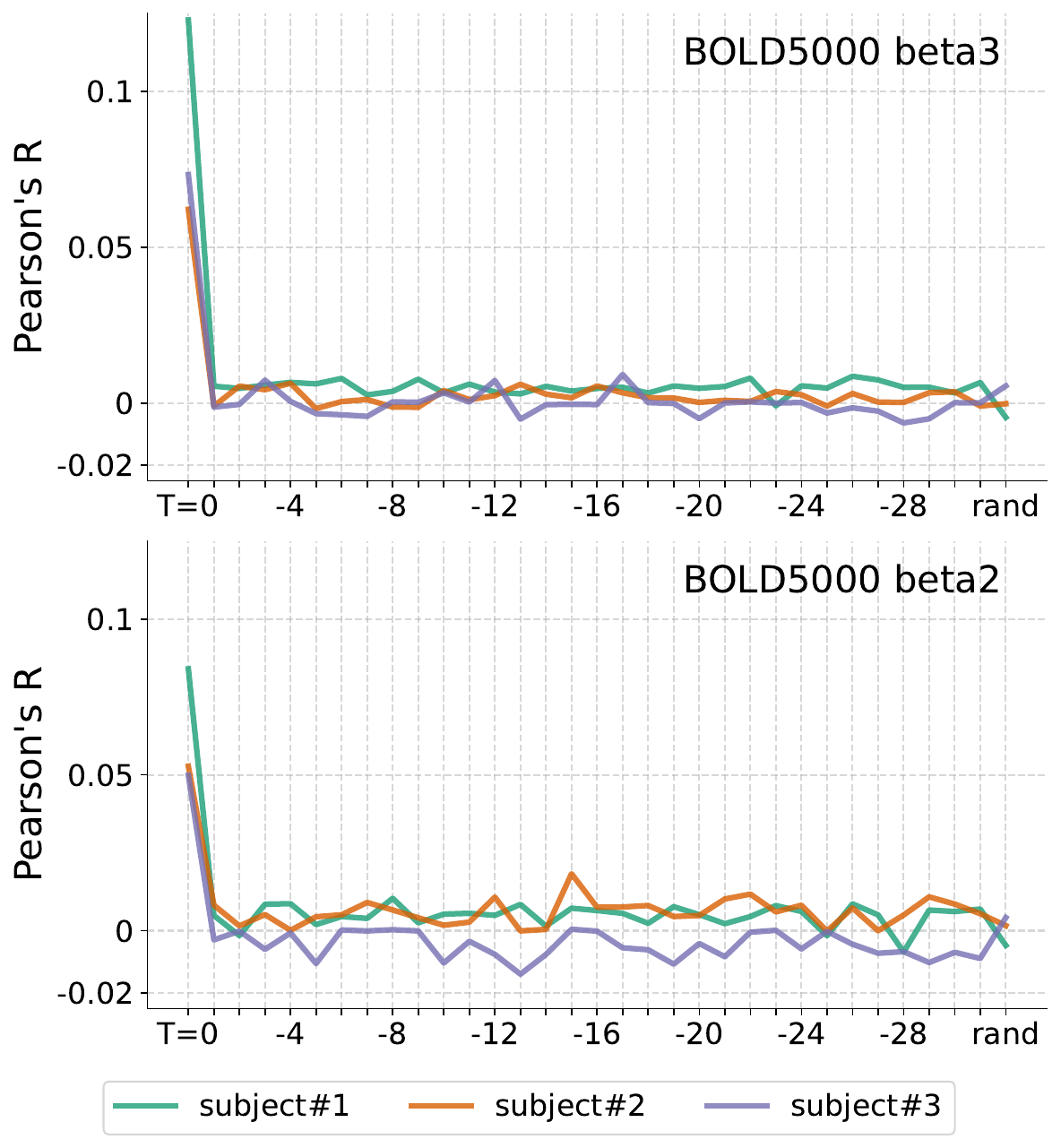}
    \caption{\textit{Replay is not observed in BOLD5000}. Results are showed as single-trail Pearson's $\mathbf{r}$, averaged over all visual brain voxels.}
    \label{fig:bold5000}
\end{figure}

\begin{figure*}[t]
    \centering

    \begin{subfigure}[b]{\textwidth}
        \includegraphics[width=\textwidth]{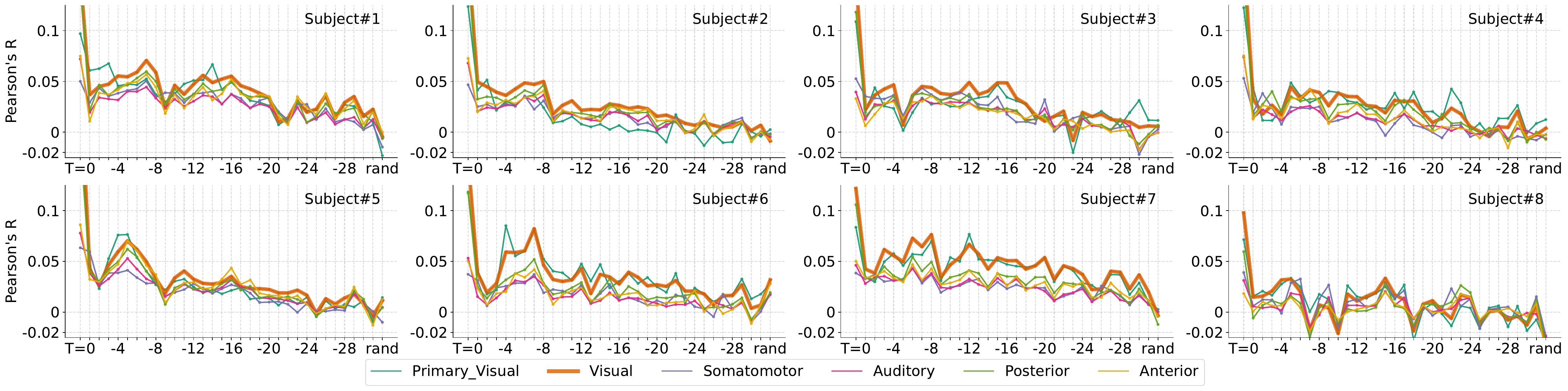}
        \caption{Cerebral Cortex, beta3}
        \label{fig:prevb3a}
    \end{subfigure}

    \begin{subfigure}[b]{\textwidth}
        \includegraphics[width=\textwidth]{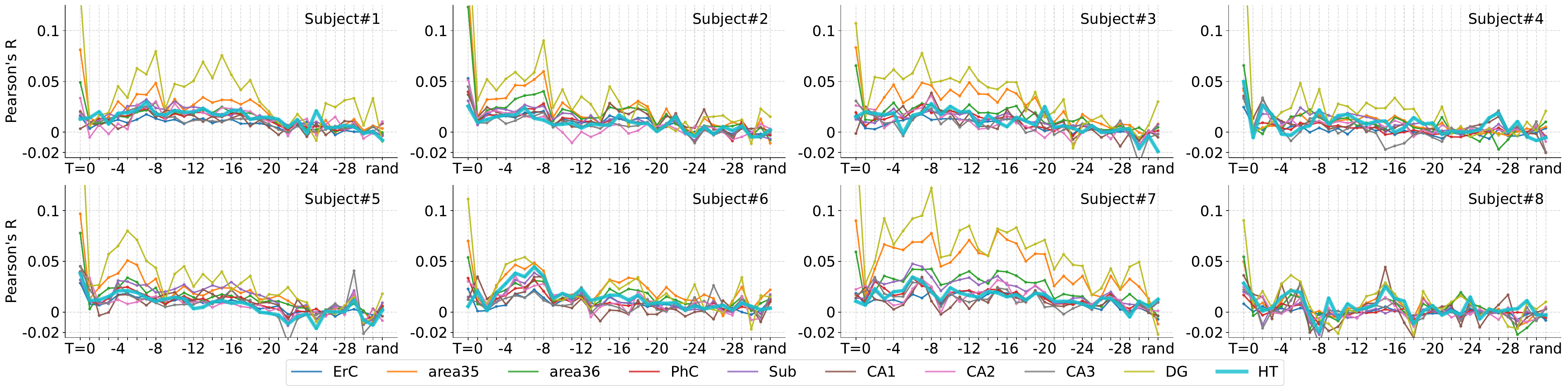}
        \caption{Hippocampus, beta3}
        \label{fig:prevb3b}
    \end{subfigure}    

    \begin{subfigure}[b]{\textwidth}
        \includegraphics[width=\textwidth]{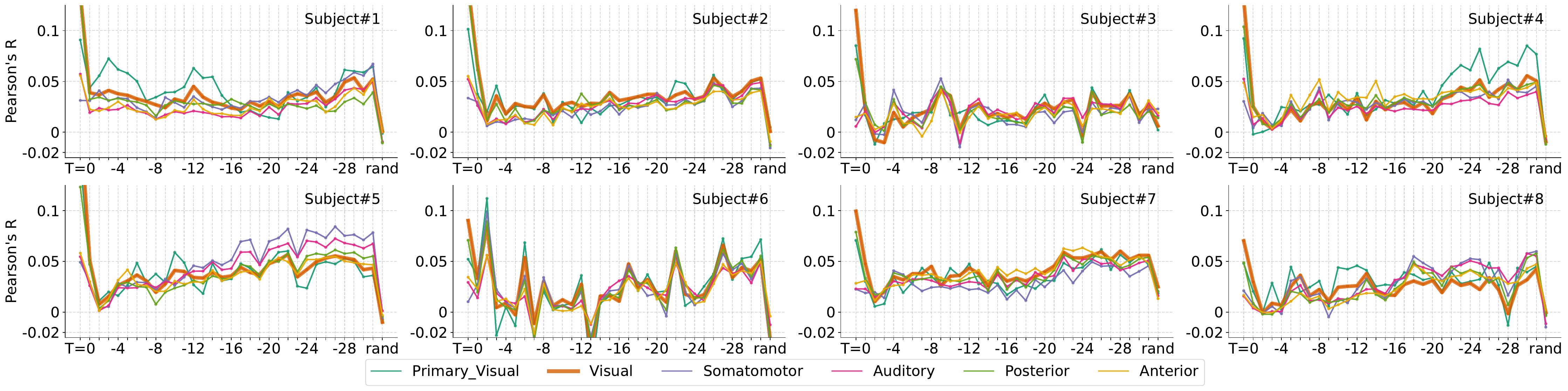}
        \caption{Cerebral Cortex, beta2}
        \label{fig:prevb2a}
    \end{subfigure}

    \begin{subfigure}[b]{\textwidth}
        \includegraphics[width=\textwidth]{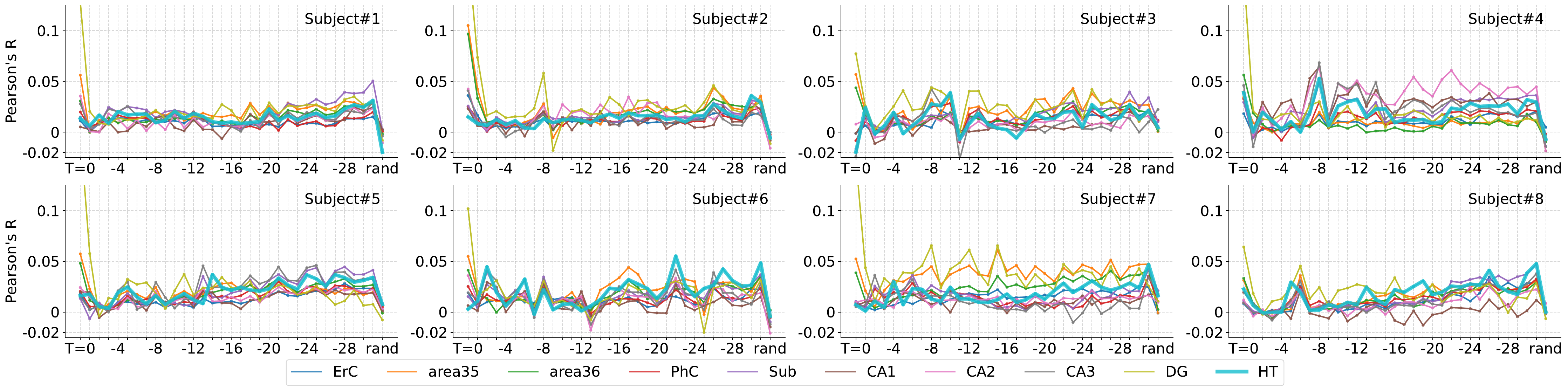}
        \caption{Hippocampus, beta2}
        \label{fig:prevb2b}
    \end{subfigure} 
    
\caption{\textbf{Periodic delayed response of previously seen images}. The x-axis is the index of the presented image, y-axis is prediction score averaged in an ROI. Models are trained using only one image frame as input and whole brain as output, a total of $32 \times 8 \times 2$ models are trained. Short-wave replay (period=6) can be observed in beta3 (a) (b), long-wave replay (period=30) can be observed in beta2 (c) (d). Note: beta3 removes T=0 unrelated signal as 'noise'.}
\label{fig:prevb3}
\end{figure*}

\paragraph{Periodic delayed response observation} 
It's worth noting that each trial is only 4 sec, the hemodynamic response function (HRF) lasts for 30 sec, and there is a significant overlap in adjacent trials. In the HRF overlap case, we expect to find information related to the previous frame to be monotonic decreasing. However, we found it periodically osculating in periods of 6 to 8 (Figure \ref{fig:replay_cortex} \ref{fig:prevb3a}).  Furthermore, we found T=-6 more predicted information in hippocampus tail than T=0 (Figure \ref{fig:replay_hippocampus} \ref{fig:prevb3b}). This suggests a read operation is done at T=-6. On the other hand, results in beta2 preparation show a peak at T=-24 to T=-30 (Figure \ref{fig:prevb2a}). T=-30 frame is presented 120 sec ago. Replay is not observed in BOLD5000 (Figure \ref{fig:bold5000}), invalidating the concern that an error in data preprocessing causes replay.

\begin{figure}
    \centering
    \includegraphics[width=0.45\textwidth]{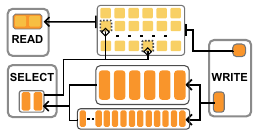}
    \caption{\textit{Hypothetical hippocampus read-write operation.} Write operation goes to both stored memory (yellow) and queues (orange). Hippocampus maintain queues of multiple size, items popped out from queue is selected for read (dashed square).}
    \label{fig:hippocampus_theory}
\end{figure}

\paragraph{Memory replay theory}
Our hypothesis is the following.  The hippocampus maintains memory queues (Figure \ref{fig:hippocampus_theory}), meant to mirror and track information stored in working memory (Figure \ref{fig:theory}). As information in working memory is overwritten, the hippocampus reads the mirrored memory and sends it to the cortex. Hippocampus maintains both short queues and long queue. Memory is being read from both short and long queues at each time point. In the cortex, image-related memory is generated in the whole brain, distributed across visual, auditory, and motor brain. Current and memory image latent codes are fused and saved to both working memory and hippocampus. This hypothesized mechanism would: 1) increase the amount of information stored in working memory and 2) enhance memory by reciting. 

\begin{table}[t]
\centering
\small 
\begin{tblr}{
  cell{1}{2} = {c=4}{},
  cell{1}{6} = {c=3}{},
  cell{2}{3} = {c=3}{},
  cell{2}{6} = {c=2}{},
  hline{1,11} = {-}{0.08em},
  hline{4} = {-}{0.05em},
  colsep = 3.5pt, 
}
\textbf{Method} & \textbf{Input}  &                    &                 &               & \textbf{Score}                      &                                       &                                    \\
                & \textbf{Images} & \textbf{Condition} &                 &               & \textbf{Private}                    &                                       & \textbf{Public}                    \\
\textbf{Model } & \textbf{Frame}  & \textbf{M}    & \textbf{B} & \textbf{T} & {\textbf{$\mathbf{r}$}\\\textbf{$\pm 0.005$}} & {\textbf{$\mathbf{r^2}$}\\\textbf{$\pm 0.005$}} & {\textbf{score}\\\textbf{$\pm 1$}} \\
Full Mem        & T=-32:0         & \checkmark             & \checkmark          & \checkmark        & 0.398                               & 0.180                                 & \textbf{66.80}                              \\
w/o time        & T=-32:0         & \checkmark             & \checkmark          &               & 0.400                               & 0.182                                 & 66.09                              \\
w/o answer      & T=-32:0         & \checkmark             &                 & \checkmark        & 0.392                               & 0.176                                 & 64.92                              \\
w/o memC        & T=-32:0         &                    & \checkmark          & \checkmark        & 0.394                               & 0.177                                 & 67.02                              \\
w/o memI        & T=0             & \checkmark             & \checkmark          & \checkmark        & 0.363                               & 0.155                                 & 62.65                              \\
w/o memIC       & T=0             &                    & \checkmark          & \checkmark        & 0.363                               & 0.156                                 & 62.24                              \\
Baseline        & T=0             &                    &                 &               & 0.332                               & 0.135                                 & \textbf{58.82}                              
\end{tblr}
\captionsetup{font=normalsize} 
\caption{Model with various input on the NSD dataset. For images, $T=-32:0$ is all previous 32 frames, $T=0$ is the current frame. For Conditions, inputs are engineered from experimental design and behavior response, \textbf{M}: memory related flags and statistics, \textbf{B}: behavior response, \textbf{T}: relative time to the scan session. The same model is trained where deactivated inputs are masked with zeros. Models are trained with the help of distillation from the best ensemble model (public score 70.85). Standard deviation across random seeds is reported in the top row. Private score is single-trial, public score is repetition-averaged.}
\label{tab:input}
\end{table}

\begin{table}
\centering
\begin{tblr}{
  cell{1}{2} = {c=4}{},
  hline{1,6} = {-}{0.08em},
  hline{3} = {-}{0.05em},
}
                & \textbf{ROI score ($\pm 0.005)$}     &                 &                   &                   \\
\textbf{Method} & {\textbf{Primary }\\\textbf{Visual}} & \textbf{Visual} & \textbf{Auditory} & \textbf{Anterior} \\
Full            & 0.300                                & 0.344           & 0.188             & 0.248             \\
w/o RM          & 0.243                                & 0.318           & 0.191             & 0.256             \\
w/o LS          & 0.287                                & 0.333           & 0.184             & 0.242             
\end{tblr}
\captionsetup{font=normalsize} 
\caption{Ablation study on \texttt{RetinaMapper} and \texttt{LayerSelector}. Whole-brain models are trained with all inputs including previous frames and condition vectors. \texttt{w/o RetinaMapper} drops the branch from feature summation. \texttt{w/o LayerSelector} replace it with simply average all layers. Standard deviation across random seeds are reported in the top row. Score is single-trail Pearson's $\mathbf{r}$.}
\label{tab:model}
\end{table}

\begin{table}[t]
\centering
\small 
\begin{tblr}{
  cell{1}{1} = {c=4}{},
  cell{1}{5} = {c=3}{},
  cell{2}{1} = {c=4}{},
  cell{2}{5} = {c=2}{},
  hline{1,12} = {-}{0.08em},
  hline{4} = {-}{0.05em},
  colsep = 3pt, 
}
\textbf{Method} &                 &             &                                      & \textbf{Score}   &                         &                 \\
                &                 &             &                                      & \textbf{Private} &                         & \textbf{Public} \\
\textbf{Model } & \textbf{Recipe} & \textbf{ID} & {\textbf{Distill}\\\textbf{Teacher}} & {\textbf{$\mathbf{r}$}\\\textbf{$\pm 0.005$}} & {\textbf{$\mathbf{r^2}$}\\\textbf{$\pm 0.005$}} & {\textbf{score}\\\textbf{$\pm 1$}} \\
Full Mem        & ensemble        & 1           &                                      & 0.416            & 0.195                   & \textbf{70.85}  \\
Full Mem        & naive           & 2           &                                      & 0.385            & 0.170                   & 63.46           \\
Full Mem        & naive           &             & 1                                    & 0.398            & 0.180                   & \uline{66.80}   \\
Baseline        & naive           & 3           &                                      & 0.324            & 0.130                   & 55.07           \\
Baseline        & naive           &             & 3                                    & 0.333            & 0.136                   & 58.22           \\
Baseline        & naive           &             & 2                                    & 0.326            & 0.131                   & 55.94           \\
Baseline        & naive           &             & 1                                    & 0.332            & 0.134                   & \uline{58.82}   \\
Baseline        & ensemble        &             & 1                                    & 0.341            & 0.141                   & \textbf{61.39}  
\end{tblr}
\captionsetup{font=normalsize} 
\caption{Models trained with naive and ensemble recipes on the NSD dataset. Full Mem model is trained with all inputs, baseline model is trained with only current frame input. Naive recipe is training one all-voxel model, ensemble recipe is training multiple ROI models with different atlas configurations. Distillation is to mimic the final output of the teacher model, teacher model is marked with \textit{ID}. Standard deviation across random seeds are reported in the top row. Private score is single-trial, public score is repetition-averaged. Bold marks best ensemble model, underline marks best single model.}
\label{tab:recipe}
\end{table}

\section{Limitations}
\label{sec:limitations}
\begin{figure*}[t]
    \centering
    \includegraphics[width=\textwidth]{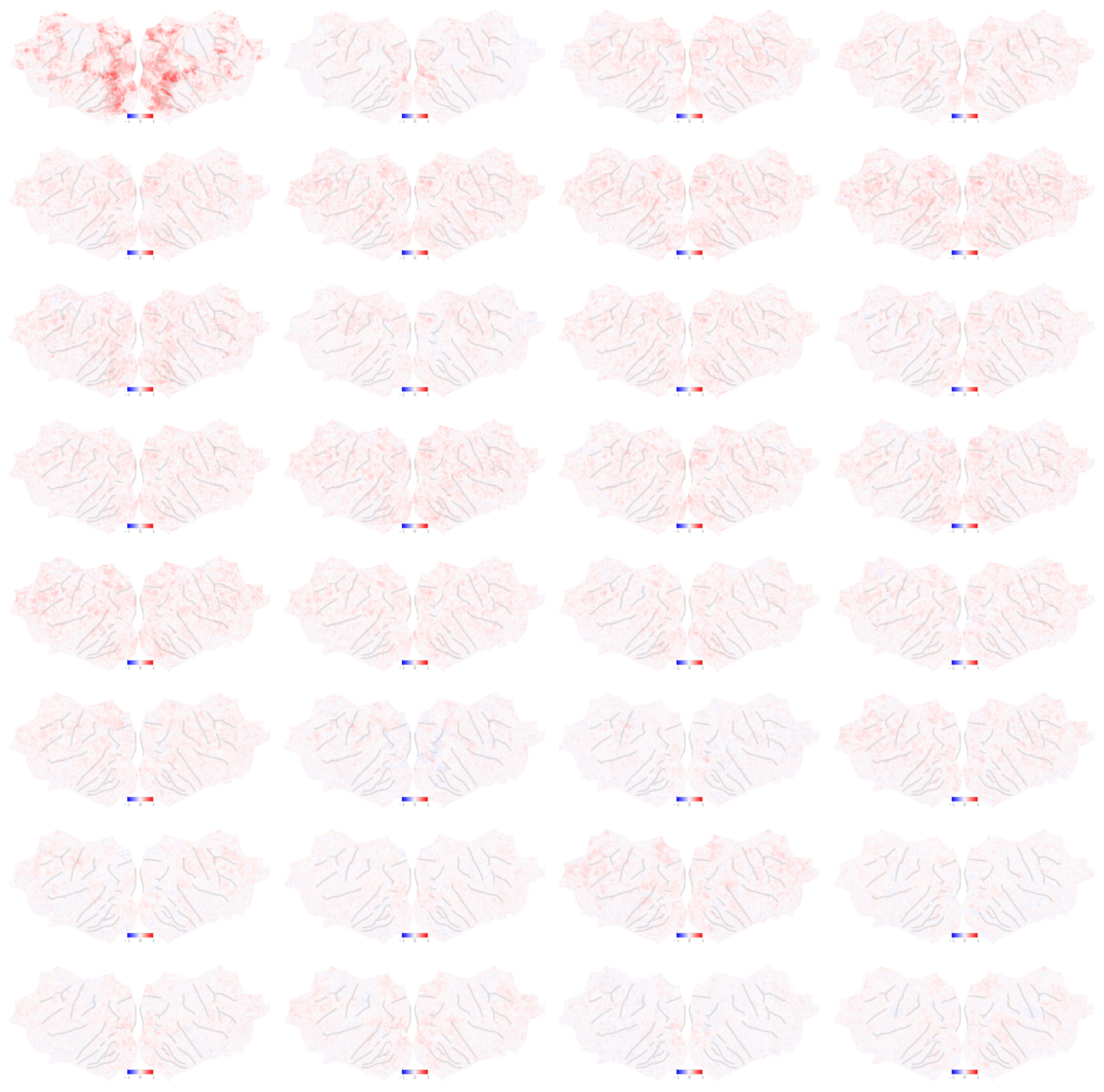}
    \caption{Prediction score with input limited to each previous frame, subject\#1 \textbf{beta3}. Input frame from T=0 to T=-31 is placed from top left to bottom right. For other subjects please see online video.}
    \label{fig:prev4x8b3}
\end{figure*}

\begin{figure*}[t]
    \centering
    \includegraphics[width=\textwidth]{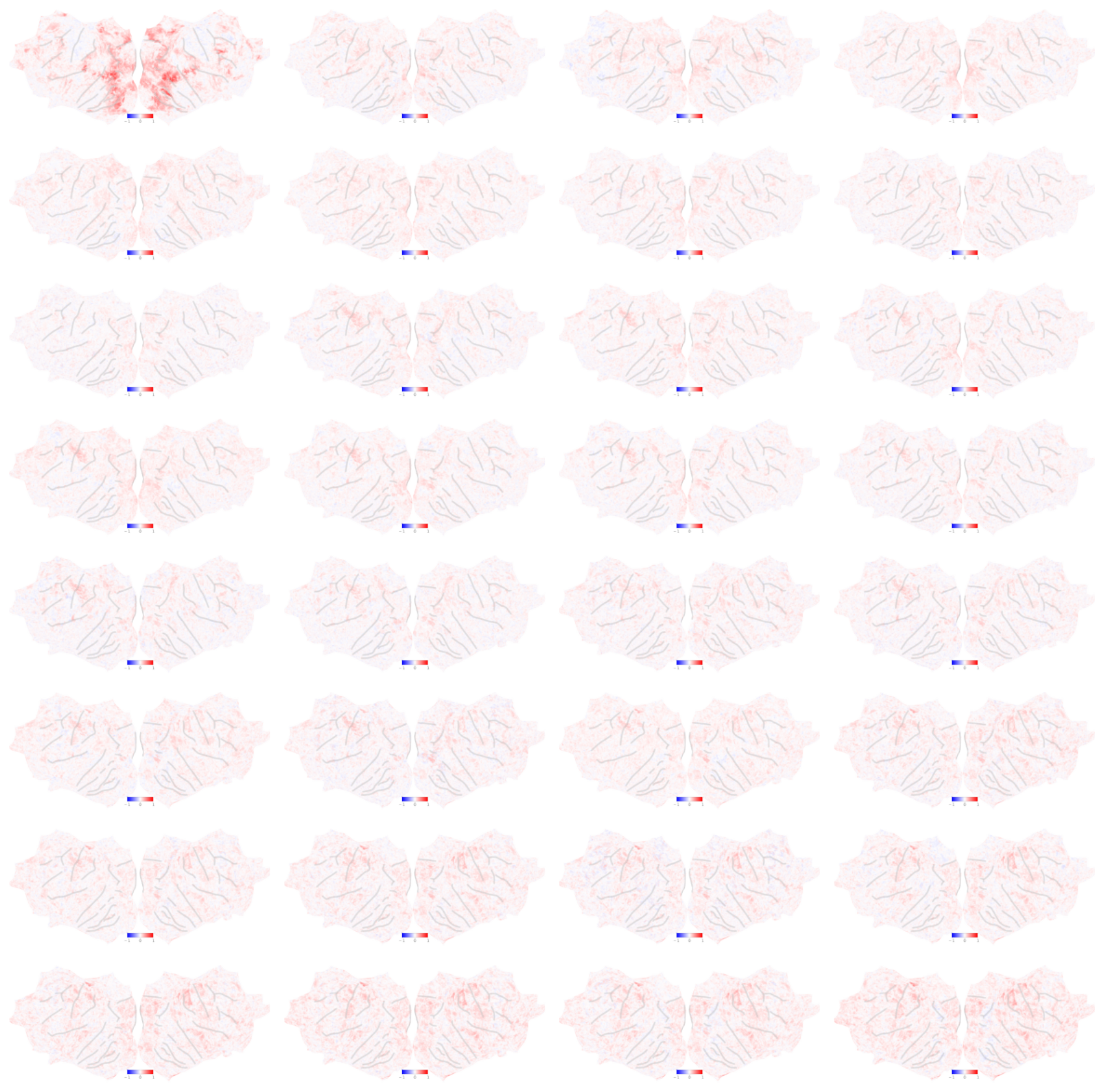}
    \caption{Prediction score with input limited to each previous frame, subject\#1 \textbf{beta2}. Input frame from T=0 to T=-31 is placed from top left to bottom right. For other subjects please see online video.}
    \label{fig:prev4x8b2}
\end{figure*}

\paragraph{Tracker does not track neuron firing pattern.} 
The prediction target for our model is GLM model weight for HRF. This will treat a constantly low and high BOLD signal both as "no signal".  Thus our ``information tracker" can only track the change of BOLD signal in the HRF time-scale, but fail to track memory stored as a neuron firing pattern that stays constant for a long time. This might be the reason for the contradiction that: our theory shows T=0 to T=-6 information is stored in working memory, but we did not find a constantly high prediction score for T=0 to T=-6 anywhere in the cortex (Figure \ref{fig:prev4x8b3} \ref{fig:prev4x8b2}).


\paragraph{Retinotopy module is resource limited.} 
As an engineering compromise, our \texttt{RetinaMapper} sample only one point from the latent image grid, thus unable to replicate receptive field size change. \texttt{LayerSelector} might mitigate this concern. However, the overall Retinotopy module has less degree-of-freedom than fitting a Gabor filter for each voxel \cite{allen_massive_2022}. 

\paragraph{Gap between theory and model.}
It's an outstanding challenge to train a model that runs forward in time only as in the theory \cite{cheng_xmem_2022}. As an engineering compromise, our Memory Encoding Model takes the previous 32 frames as input. Thus, our model is unable to capture longer-term memory. At the time of writing, we are still determining if this will compromise model performance. Furthermore, our computational model reconstructs the whole memory buffer from scratch on each forward pass.


\newpage
\section{Future Work}
\label{sec:future_work}
\paragraph{Denoising} We demonstrated results that the current denoising pipeline removes memory-related signal (Table \ref{tab:app_b23}). The issue stems from using PCA on low noise ceiling voxels to derive noise regressor. Future work might consider a new way for noise regressor. It also incentivizes removing time-related signals (Figure \ref{fig:c3a}).

\paragraph{Validate periodic delayed response.} To our knowledge, the exact cause of periodic delayed response remains unknown. We suspect it to be: 1) hard to find on noisy data. 2) related to long-term memory tasks. 3) related to large working memory tasks.

\paragraph{A note on data capturing.} The brain is a complex machine. Any source of information could lead to a better brain encoding model, even if the information is not perfect, such as un-calibrated eye tracking camera \cite{lurz_generalization_2021} or eye voxels \cite{frey_magnetic_2021}.

\newpage
\section{Conclusions}
\label{sec:conclusions}
We proposed a theory for a memory process that impacts the brain encoding model. The theory is based on experiment observation of periodic delayed response and hippocampus tail activity. We found the whole brain during the vision-memory task is largely predictable thanks to memory-related information.

\paragraph{Acknowledgement}
We thank Shi Gu, Yuanning Li for help review this paper, Yingtian Tang for insightful discussions. Cortex plot are made with \texttt{pycortex} \cite{gao_pycortex_2015}. Models are build with \texttt{pytorch lightning} \cite{falcon_pytorch_2019}.

{\small
\bibliographystyle{ieee_fullname}
\bibliography{main}
}


\clearpage
\pagestyle{fancy}
\fancyhead{}
\fancyhead[RO,LE]{\textbf{Periodic delayed response on all ROIs}}

\clearpage
\pagestyle{fancy}
\fancyhead{}
\fancyhead[RO,LE]{\textbf{Comparison of beta3 and beta2}}

\begin{table*}[t]
\centering
\begin{subtable}{0.9\textwidth}
    \centering
    \begin{tblr}{
      cell{1}{4} = {c=5}{},
      hline{1,13} = {-}{0.08em},
      hline{2} = {4-8}{0.03em},
      hline{3,7,11} = {-}{},
      colsep = 6.5pt, 
    }
                  &                &                   & \textbf{ROI score (Cerebral Cortex) } &                   &                    &                   &                      \\
    \textbf{Beta} & \textbf{Input} & \textbf{whole-brain score($r$)} & \textbf{Visual}                 & \textbf{Auditory} & \textbf{Posterior} & \textbf{Anterior} & \textbf{Somatomotor} \\
    b3            & T=-32:0        & 0.174             & 0.244                           & 0.144             & 0.220              & 0.180             & 0.162                \\
    b3            & T=0            & 0.079             & 0.158                           & 0.058             & 0.131              & 0.058             & 0.048                \\
    b3            & T=-6           & 0.036             & 0.049                           & 0.032             & 0.040              & 0.037             & 0.034                \\
    b3            & T=-28          & 0.007             & 0.014                           & 0.005             & 0.007              & 0.006             & 0.006                \\
    b2            & T=-32:0        & 0.237             & 0.302                           & 0.211             & 0.272              & 0.211             & 0.265                \\
    b2            & T=0            & 0.055             & 0.123                           & 0.036             & 0.096              & 0.039             & 0.026                \\
    b2            & T=-6           & 0.022             & 0.031                           & 0.019             & 0.021              & 0.021             & 0.024                \\
    b2            & T=-28          & 0.037             & 0.038                           & 0.039             & 0.035              & 0.036             & 0.041                \\
    b3            & T=rand         & -0.002            & 0.001                           & -0.002            & -0.003             & -0.002            & -0.004               \\
    b2            & T=rand         & -0.002            & -0.002                          & -0.002            & -0.005             & 0.002             & -0.005               
    \end{tblr}
\end{subtable}
\hfill
\begin{subtable}{0.9\textwidth}
    \centering
    \begin{tblr}{
      cell{1}{3} = {c=10}{},
      hline{1,13} = {-}{0.08em},
      hline{2,3} = {3-12}{},
      hline{7,11} = {-}{0.05em},
    }
                  &                & \textbf{ROI score (Medial Temporal Lobe)} &                              &                              &              &              &              &              &              &             &             \\
    \textbf{Beta} & \textbf{Input} & \textbf{ErC}                        & {\textbf{area}\\\textbf{35}} & {\textbf{area}\\\textbf{36}} & \textbf{PhC} & \textbf{Sub} & \textbf{CA1} & \textbf{CA2} & \textbf{CA3} & \textbf{DG} & \textbf{HT} \\
    b3            & T=-32:0        & 0.058                               & 0.125                        & 0.105                        & 0.077        & 0.080        & 0.079        & 0.100        & 0.095        & 0.210       & 0.075       \\
    b3            & T=0            & 0.024                               & 0.081                        & 0.070                        & 0.029        & 0.022        & 0.024        & 0.035        & 0.028        & 0.150       & 0.022       \\
    b3            & T=-6           & 0.013                               & 0.037                        & 0.025                        & 0.020        & 0.024        & 0.017        & 0.016        & 0.017        & 0.057       & 0.022       \\
    b3            & T=-28          & 0.001                               & 0.006                        & 0.004                        & 0.001        & 0.001        & 0.002        & 0.001        & 0.002        & 0.008       & 0.003       \\
    b2            & T=-32:0        & 0.109                               & 0.127                        & 0.114                        & 0.131        & 0.140        & 0.131        & 0.147        & 0.170        & 0.190       & 0.156       \\
    b2            & T=0            & 0.015                               & 0.055                        & 0.050                        & 0.017        & 0.014        & 0.016        & 0.026        & 0.019        & 0.125       & 0.012       \\
    b2            & T=-6           & 0.009                               & 0.017                        & 0.012                        & 0.014        & 0.017        & 0.010        & 0.011        & 0.017        & 0.023       & 0.016       \\
    b2            & T=-28          & 0.019                               & 0.025                        & 0.025                        & 0.019        & 0.031        & 0.015        & 0.019        & 0.019        & 0.019       & 0.024       \\
    b3            & T=rand         & -0.000                              & 0.003                        & 0.002                        & 0.002        & 0.002        & -0.002       & 0.001        & 0.001        & 0.010       & -0.002      \\
    b2            & T=rand         & 0.002                               & -0.002                       & -0.001                       & 0.002        & 0.001        & -0.003       & -0.005       & 0.001        & -0.004      & -0.002      
    \end{tblr}
\end{subtable}
\caption{Single-trial Pearson's $\mathbf{r}$, each number is mean score over voxels from all subjects. Identical whole-brain models are trained with varying input image frame(s), each row is one model, results are for subject\#1. $T=0$ is current frame, $T=-n$ is previous \#n frame, time between two frames is 4s. Beta is denosing pipeline variants, beta version3 removes $T=0$ unrelated signal as 'noise'. Compared to b2, b3 has higher score at $T=0$ but lower score on $T=-32:0$}
\label{tab:app_b23}
\end{table*}

\clearpage
\pagestyle{fancy}
\fancyhead{}
\fancyhead[RO,LE]{\textbf{Periodic delayed response cortex plot, beta3}}

\clearpage
\pagestyle{fancy}
\fancyhead{}
\fancyhead[RO,LE]{\textbf{Periodic delayed response cortex plot, beta2}}

\end{document}